\documentclass[lettersize,journal]{IEEEtran}
\usepackage{amsmath,amsfonts}
\usepackage{algorithmic}
\usepackage{algorithm}
\usepackage{array}
\usepackage{subfigure} 
\usepackage{textcomp}
\usepackage{stfloats}
\usepackage{url}
\usepackage{verbatim}
\usepackage{graphicx}
\usepackage{caption} 
\usepackage{cite}
\usepackage{float} 
\usepackage{booktabs}
\usepackage{threeparttable}    
\usepackage{multirow}
\usepackage{hyperref}
\makeatletter

\newcommand{\Rmnum}[1]{\expandafter\@slowromancap\romannumeral #1@}
\makeatother

\begin{document}
%


\title{A Sparse Cross Attention-based Graph Convolution Network with Auxiliary Information Awareness for Traffic Flow  Prediction}

\author{Lingqiang Chen,
        Qinglin Zhao, ~\IEEEmembership{Senior Member, ~IEEE,}
        Guanghui Li\IEEEauthorrefmark{1},
        Mengchu Zhou, ~\IEEEmembership{Fellow, ~IEEE,}
        Chenglong Dai, and
        Yiming Feng
\thanks{Lingqiang Chen is with the School of Information and Electrical Engineering, Hebei University of Engineering, Handan 056038, China (e-mails:chenlingqiang@hebeu.edu.cn).

Guanghui Li, Chenglong Dai, and Yiming Feng are with the School of Artificial Intelligence and Computer Science, Jiangnan University, Wuxi 214122, China (e-mails: ghli@jiangnan.edu.cn; chenglongdai@jiangnan.edu.cn; ricardo\_fym@outlook.com).

Qinglin Zhao is with the School of Computer Science and Engineering, Macau University of Science and Technology, Avenida Wei Long, Taipa, Macau 999078, China (e-mail: qlzhao@must.edu.mo). 

Mengchu Zhou is with the Department of Electrical and Computer Engineering, New Jersey Institute of Technology, Newark, NJ 07102, USA (e-mail: zhou@njit.edu). \\}
\thanks{Guanghui Li is the corresponding author.}}

\markboth{Journal of \LaTeX\ Class Files,~Vol.~*, No.~*, *~*}%
{Shell \MakeLowercase{\textit{et al.}}: Bare Demo of IEEEtran.cls for Computer Society Journals}

\maketitle

\begin{abstract}
	Deep graph convolution networks (GCNs) have recently shown excellent 
	performance in traffic prediction tasks. However, they face some challenges. 
	First, few existing models consider the influence of auxiliary information, 
	i.e., weather and holidays, which may result in a poor grasp of 
	spatial-temporal dynamics of traffic data. Second, both the construction of a 
	dynamic adjacent matrix and regular graph convolution operations have 
	quadratic computation complexity, which restricts the scalability of 
	GCN-based models. To address such challenges, this work proposes a  deep 
	encoder-decoder model entitled AIMSAN. It contains an \underline{a}uxiliary 
	\underline{i}nformation-aware module (AIM) and \underline{s}parse cross 
	\underline{a}ttention-based graph convolution network (SAN). The former 
	learns multi-attribute auxiliary information and obtains its embedded 
	presentation of different time-window sizes. The latter uses a 
	cross-attention mechanism to construct dynamic adjacent matrices by fusing 
	traffic data and embedded auxiliary data. Then, SAN applies diffusion GCN on 
	traffic data to mine rich spatial-temporal dynamics. Furthermore, AIMSAN 
	considers and uses the spatial sparseness of traffic nodes to reduce the 
	quadratic computation complexity. Experimental results on three public 
	traffic datasets demonstrate that the proposed method outperforms other 
	counterparts in terms of various performance indices. Specifically, the 
	proposed method has competitive performance with the state-of-the-art 
	algorithms but saves 35.74\% of GPU memory usage, 42.25\% of training time, 
	and 45.51\% of validation time on average.    
\end{abstract}


\section{Introduction}\label{sec:introduction}

\IEEEPARstart{T}{he} increasing use of vehicles makes the city's 
transportation-related problems increasingly severe, like traffic jams and air 
pollution \cite{kumar2021applications,jiang2021graph}. Therefore, research on 
traffic flow has attracted wide attention, among which traffic data prediction is 
one hot spot. However, traffic flow has complex spatial-temporal dynamics 
influenced by human activities and other factors (like weather). For example, due 
to people commuting to work, the traffic flow in the morning or evening rush hour 
is significantly higher than that in other periods on the same day. The traffic 
speed during foggy days is slower than that during clear days. Besides, those 
road repairs or accidental traffic events cause unexpected traffic jams on 
certain roads. 

In recent years, deep learning models have been widely used to solve traffic 
prediction tasks thanks to their remarkable representational ability, among which 
spatial-temporal models are the most promising ones 
\cite{ren2020global,pan2020spatio}.
In the early research stage, dividing the distributions of traffic states into 
grids on a map, the studies in \cite{zhang2017deep,yao2019revisiting} apply 
convolution neural networks (CNN) to extract spatial interactions and adopt 
recurrent neural networks (RNN) to learn temporal dependency.
However, considering the natural graph structure of traffic sensors, more 
researchers have found that it is more reasonable to mine the spatial 
interactions of traffic data in the form of graph diffusion 
\cite{li2018diffusion,ijcai2019-264}. 
Given the traffic data by a graph structure, Fig. \ref{fig:gcninonelayer} shows 
the traffic states update process after one graph convolution network (GCN) 
operation, where \begin{small}$X_{\text{n1}}$\end{small} is the current data 
state of node 1 (\begin{small}$n_1$\end{small}), 
\begin{small}$X_{\text{n1}}'$\end{small} is the updated state, and 
\begin{small}$w_{i,j}$\end{small} is the weight between 
\begin{small}$n_i$\end{small} and \begin{small}$n_j$\end{small}. In GCN, each 
node updates local traffic states by weighting the data from itself and its 
neighbors. 

\begin{small}
	\begin{figure*}[!htpb]
		\centering
		\includegraphics[width=0.9\linewidth]{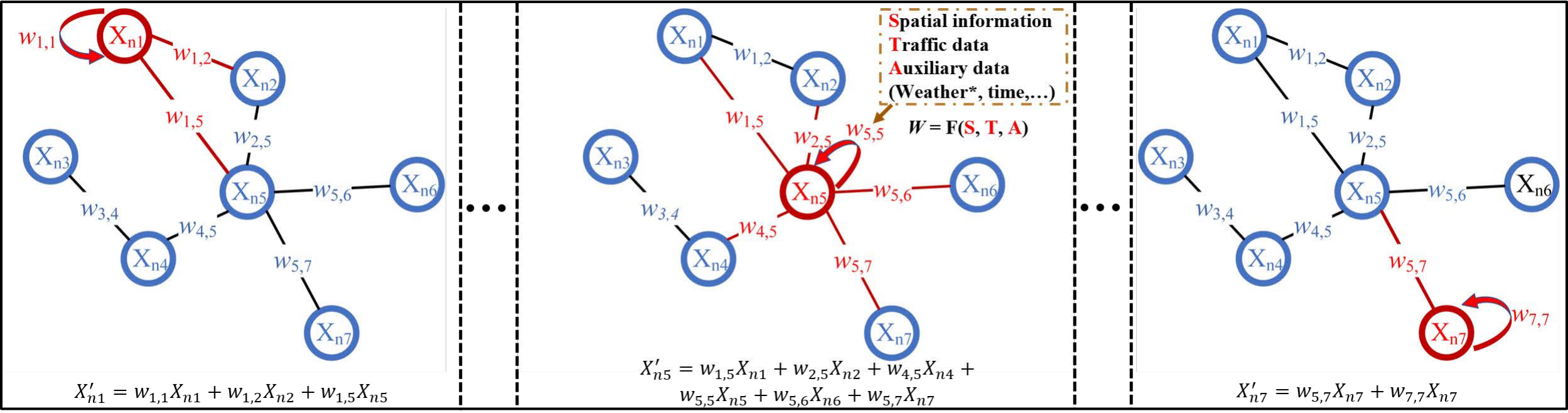} 
		\caption{Traffic states update process under one GCN operation.
		}
		\label{fig:gcninonelayer} 
	\end{figure*}
\end{small}

Recently, most researchers have focused on graph convolution, especially dynamic 
graph convolution. Those dynamic methods mainly apply specific mechanisms to 
calculate variable weights of adjacency matrices in each GCN process and then use 
the new adjacency matrices to update the traffic states.
For example, Han et al. in \cite{DMSTGCN} design a dynamic graph constructor to 
generate a time-specific adjacency tensor. As a result, those GCN models can 
learn more hidden spatial-temporal information with dynamic graph convolution 
operation. 

However, two key issues remain to be investigated for GCN-based models.
First, traffic data contain complex spatial-temporal dynamics influenced by 
traffic attributes and other factors like weather and holidays. Yet, many graph 
convolution models adopt the same graph for information extraction under all 
conditions, whether using static graphs (based on distance) or self-adaptive ones 
(based on learning), which ignores the spatial-temporal diversity of traffic data 
\cite{pan2020spatio}. Even for those dynamic GCN models 
\cite{li2021spatial,huang2022dynamical}, few existing studies consider the 
influence of auxiliary information. Those auxiliary information are 
non-negligible because they can help the model learn more about hidden traffic 
states. Therefore, in GCN operations shown in Fig. \ref{fig:gcninonelayer}, 
AIMSAN considers designing a fusion function F(S, T, A) to embed 
\underline{s}patial information, \underline{t}raffic data, and 
\underline{a}uxiliary information (like weather and holiday) for constructing an 
auxiliary-information-aware dynamic adjacent matrix, which can be used to mine 
rich spatial-temporal dynamics.
Second, with the expansion of a city, the scalability of GCN-based models is 
restricted by the traffic node size (\begin{small}$N$\end{small}) due to the 
quadratic computation complexity of regular graph convolution operations. 

This paper presents a deep encoder-decoder model called AIMSAN containing Dilated 
Causal Convolution network (DCC), AIM and SAN modules to address the above two 
issues. DCC is a temporal convolutional network that can reduce the stacking 
layer size by dilation convolutions. AIM can embed multiple-attribute auxiliary 
information in the historical or future periods into one tensor representation. 
The historical embedded data are fed into SAN, and future embedded data are added 
to the decoder part to increase the model's awareness of future auxiliary 
information. In each encoder layer, AIMSAN concatenates feed-forward data with 
the embedded data to calculate the dynamic adjacent matrix using a 
cross-attention mechanism. Subsequently, it uses dynamic adjacent matrices to 
mine rich spatial-temporal information from traffic data. In short, this work 
intends to make the following novel contributions to the field of traffic 
prediction: 

1) It proposes an auxiliary information-aware method to embed multiple attributes 
like time, position, and weather conditions and learn more about hidden traffic 
states. In the encoder layer, AIM aligns the historical auxiliary data with 
hidden traffic data in the temporal dimension to keep its focus on the 
information in the current time window. Besides, AIM embeds future auxiliary data 
in the decoder to improve its understanding of the predicted traffic state. 
Lastly, AIM is easy to be trimmed and extended due to its multi-branch structure.

2) It designs a novel dynamic adjacency matrix acquisition method that fuses 
traffic data and embedded data from AIM to obtain an auxiliary-information-aware 
dynamic adjacency matrix and mines rich spatial-temporal diversity for the first 
time.

3) It proposes a sparse cross-attention-based graph convolution (SAN) module that 
applies the spatial sparseness of traffic nodes to a graph convolution process, 
thus reducing the computational complexity.

4) It evaluates the proposed approach and existing state-of-the-art ones on three 
real-world datasets. Experimental results demonstrate that it outperforms its 
peers in most cases.

The rest of this paper is organized as follows: Section \Rmnum{2} introduces the 
related work about traffic flow prediction. Then, Section \Rmnum{3} gives the 
problem statement of traffic prediction. Section \Rmnum{4} describes the 
implementation of AIMSAN. Section \Rmnum{5} presents the experimental results and 
discussions. Finally, Section \Rmnum{6} concludes this work.

\section{Related work}
Deep learning methods have been widely used for traffic prediction. This section 
introduces some deep traffic prediction models with or without GCN.

\subsection{Traffic prediction without GCN}
Those traffic prediction models without GCN are mainly based on Stacked 
Autoencoder (SAE), Deep Belief Networks (DBN), Recurrent Neural Networks (RNN), 
Convolutional Neural Networks  (CNN), or hybrid ones 
\cite{tedjopurnomo2020survey}. In the early stage, some studies consider 
optimizing the computation time of training models. Therefore, they apply SAEs 
and DBNs in traffic flow prediction 
\cite{huang2014deep,lv2014traffic,soua2016big}. However, many researchers verify 
that they fail to capture the spatial or temporal aspect of the traffic data and 
thus tend to perform worse than those spatial-temporal-correlation-based neural 
networks \cite{cheng2018deeptransport}. Differently, RNN-based models are 
naturally suitable for traffic time-series data mining. For example, Fouladgar et 
al. in \cite{fouladgar2017scalable} constructed the traffic data as a matrix and 
used an LSTM to capture the hidden spatial-temporal correlation of data. However, 
with the growth of the traffic flow prediction series, those RNN-based models 
require long input sequences, which can heavily impact the training and inference 
efficiency. 

CNN is another optional deep-learning framework for traffic prediction tasks 
thanks to its ability to capture the correlation between different regions or 
time slots. Like RNNs, the temporal convolutional network (TCN) is a typical 1D 
CNN model that can be used to analyze traffic data from the temporal dimension. 
Considering the spatial-temporal attributes of traffic data, some CNN-based 
models perform 2D CNN on traffic data in both temporal and spatial dimensions 
\cite{fouladgar2017scalable}, like an image feature extraction process 
\cite{9146985,9797232}. 

In addition, hybrid models have been extensively studied because they can utilize 
the strengths of their individual components. For example, Du et al. 
\cite{du2017traffic} combine both CNN and LSTM components to capture spatial and 
temporal features, respectively, and fuse the outputs from the above two networks 
to form the final prediction. Another version of hybrid methods adds 
convolutional operation into the dense kernel of the LSTM unit such that the 
improved LSTM model can simultaneously capture the spatial-temporal correlation 
of traffic data. Though CNNs can efficiently mine the spatial patterns of traffic 
data, most of them regard the spatial relationship of traffic flow as a simple 
Euclidean structure while ignoring the complex spatial relationship among nodes. 
Thus 2D CNN is not the best choice for the prediction of graph structure traffic 
data.

\subsection{Traffic prediction with GCN}
The graph convolution network (GCN) \cite{9395542,9875203,9944200} is an 
effective method for data extraction, especially for those data with graph 
structures like traffic data. Unlike 2D CNN, a GCN model learns the 
transformation of traffic data through graph diffusion. GCN models are roughly 
divided into two groups: traditional and dynamic ones. 

Traditional GCN models apply a fixed adjacent matrix in each graph convolution 
layer \cite{yu2018spatio,li2018diffusion, ijcai2019-264}. For example, in 
\cite{ijcai2019-264}, the authors use both static and learnable adjacent 
matrices, where the weights of a static adjacent matrix are initialized according 
to the distance between each pair of traffic nodes, and the learnable adjacent 
matrix is constructed by the matrix multiplication of two embedding matrices. 
However, whether using predefined adjacent matrices (like distance-based or 
content-based adjacent matrices) or learnable ones, those adjacent matrices are 
fixed once a model is trained, thus ignoring the sample and position-specific 
characteristics of data \cite{pan2019urban}.

Dynamic GCNs are increasingly being applied for traffic prediction tasks, which 
construct variable adjacency matrices to mine the rich spatial-temporal 
information \cite{li2021spatial,huang2022dynamical}. Considering the daily 
periodicity of traffic status, the Dynamic and Multi-faceted Spatio-Temporal 
Graph Convolution Network (DMSTGCN) in \cite{DMSTGCN} uses an inverse process of 
Tucker decomposition to construct a dynamic adjacent matrix 
\begin{small}$\mathbf{A}\in\mathbb{R}^{N_t\times N\times N}$\end{small}, where 
\begin{small}$N_t$\end{small} is the number of time slots in a day. Therefore, 
DMSTGCN can learn the time-specific spatial dependencies of road segments. 
Besides, it applies a dynamic graph convolution module to aggregate the hidden 
states of neighbor nodes to focal nodes by passing messages on the dynamic 
adjacency matrices. Differently, Li et al. \cite{li2021spatial} propose a 
spatial-temporal fusion graph neural network (STFGNN). They use a data-driven 
method to generate a dynamic adjacent matrix of size \begin{small}$N \times 
N$\end{small}. By combining multiple spatial and temporal graphs, STFGNN can 
efficiently learn the spatial-temporal correlations simultaneously. However, 
stacking adjacent matrices of size \begin{small}$4N \times 4N$\end{small} in 
STFGNN  dramatically increases computation and memory costs.
In addition, attention-based graph neural networks are used to model the 
spatial-temporal dynamic of traffic data \cite{guo2021learning,comcom}. For 
example, Guo et al. \cite{guo2021learning} design a self-attention module to 
capture temporal dynamics of traffic data and use a dynamic graph convolution 
module to learn spatial correlations. Dynamic GCN models can discover more hidden 
correlations than traditional GCN models. However, they only use traffic data 
(like traffic speed or flow) and ignore the influence of auxiliary factors.

Several methods have recently utilized auxiliary information of road segments, 
like speed limit, joint angle, and points of interest, to mine the hidden feature 
of traffic data \cite{shin2020incorporating, lee2022ddp, pan2020spatio}. For 
example, Shin et al. \cite{shin2020incorporating} propose a multi-weight traffic 
graph convolutional network (MW-TGC). They conduct graph convolution operations 
on speed data with multi-weighted adjacency matrices to combine the auxiliary 
features, including speed limit, distance, and angle. In \cite{DMSTGCN}, the 
authors explore dynamic and multi-faceted spatial-temporal characteristics 
inherent in traffic data and use traffic volumes to predict future traffic speed. 
In addition, Pan et al. \cite{9096591}
propose a meta-learning-based mode named ST-MetaNet, which applies meta-knowledge 
learners to learn the edge and node attributes of traffic sensors and uses the 
learned weights to construct a dynamic adjacent matrix to mine spatial-temporal 
information. Unlike ST-MetaNet, AIMSAN uses a cross-attention strategy to 
construct a dynamic adjacent matrix, which fuses traffic state data with multiple 
attribute data like positional, temporal and weather information in historical 
and future periods.


The most related work to this paper is our previous work (TGANet) in 
\cite{comcom}. TGANet is a dynamic graph convolutional method that uses 
self-attention-based and multi-weight adjacent matrices to mine spatial-temporal 
dynamics of traffic data.
However, it has the following deficiencies. First, like many dynamic GCN methods, 
TGANet neglects the influence of auxiliary factors like weather toward the target 
attribute. Second, the multi-weight GCN operation in TGANet requires extra 
computation costs. Unlike it, AIMSAN designs AIMs for learning the sample and 
position-specific auxiliary information and constructs an auxiliary 
information-aware adjacent matrix 
to learn rich spatial-temporal dynamics. Only using the auxiliary 
information-aware adjacent matrix, AIMSAN requires less computation overhead than 
TGANet.

\section{Problem statement}
In intelligent traffic systems, traffic monitoring nodes (called nodes for short) 
are usually deployed around the city, as shown in Fig. \ref{fig:UrbanArea}. 
The distribution of such nodes forms a graph structure 
\begin{small}$G=\{\mathbf{V}, \mathbf{E}\}$\end{small} like Fig. 
\ref{fig:UrbanArea_graph}, where \begin{small}$\mathbf{V} = \{v_1, v_2, \cdots, 
v_n\}$\end{small} is the set of  nodes, and \begin{small}$\mathbf{E}$\end{small} 
is the set of edges. In addition, the adjacency matrix 
\begin{small}$\mathbf{A}=[a_{ij}] \in \mathbb{R}^{N\times N}$\end{small} records 
the connection between each pair of nodes. If nodes 
\begin{small}$v_i$\end{small} and \begin{small}$v_j$\end{small} belong 
to \begin{small}$\mathbf{V}$\end{small}, and they are connected, then 
\begin{small}$e_{ij} \in \mathbf{E}$\end{small}, and 
\begin{small}$a_{ij}=1$\end{small}. If not connected, 
\begin{small}$a_{ij}=0$\end{small}. 

\begin{small}
	\begin{figure}[!htbp]
		\centering
		\subfigure[Urban area]{
			\includegraphics[width=0.65\linewidth]{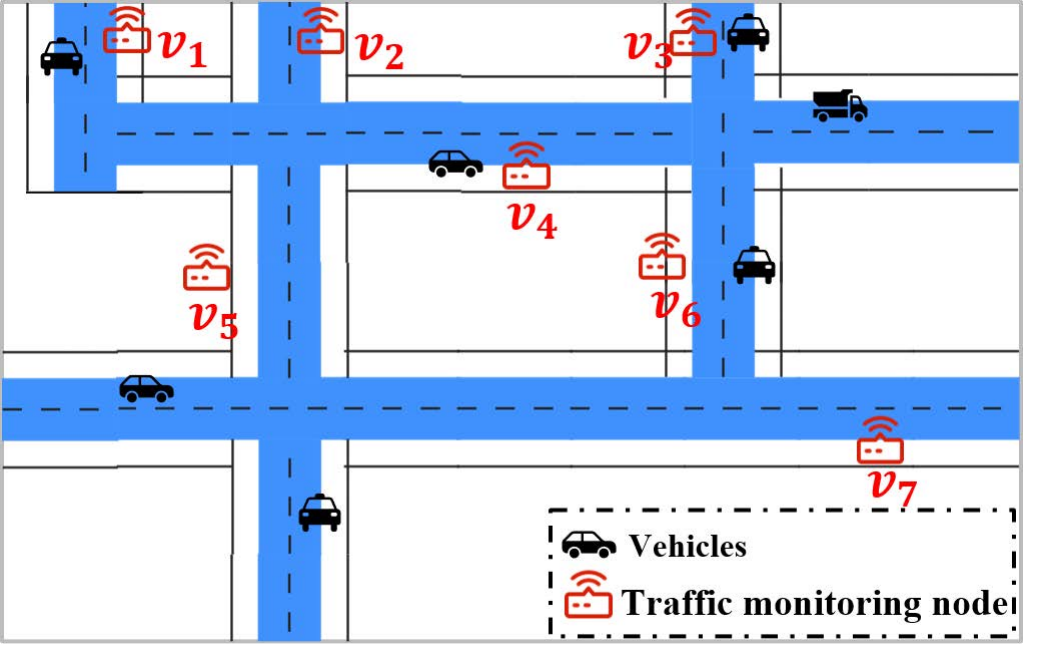}
			\label{fig:UrbanArea}
		}
		\subfigure[Graph structure]{
			\includegraphics[width=0.65\linewidth]{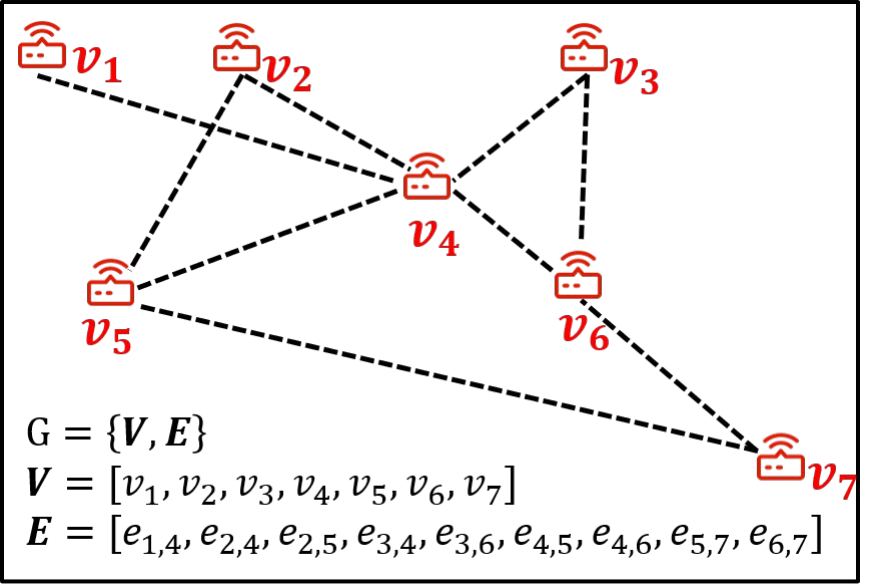}
			\label{fig:UrbanArea_graph}
		}
		\caption{An example of traffic sensor node distribution in a city.
		}
		\label{fig:distribution} 
	\end{figure}
\end{small}

\begin{table*}[!htbp] \scriptsize
	\caption{Frequently used notations}
	\label{tab:notations}
	\resizebox{\textwidth}{!}{%
		\begin{tabular}{llll}
			\toprule
			Notation & Description & Notation & Description \\
			\midrule
			G=\{$\mathbf{V}$, $\mathbf{E}$\} & \begin{tabular}[c]{@{}l@{}}A graph 
			structure, where $\mathbf{V}$ is the set of nodes, $\mathbf{E}$ is \\ 
			the set of edges\end{tabular} & $\mathbf{I}_j$/$\mathbf{S}_j$ &  
			\begin{tabular}[c]{@{}l@{}}Input data and skip data in the  
			$j^{\text{th}}$ SAN Layer\end{tabular}  \\
			\midrule
			$\mathbf{A}$/$\mathbf{A}_N$/$\mathbf{A}^b$ & 
			\begin{tabular}[c]{@{}l@{}}An adjacent matrix/normalized adjacent 
			matrix/ the \\ $b^{\operatorname{th}}$ adjacent matrix of the batch 
			adjacent matrices in \\SAN layer\end{tabular} 
			& $\mathbf{Q}$/$\mathbf{K}$/$\mathbf{V}$ & Query/Key/Value of an 
			attention module\\
			\midrule
			$\mathbb{N}$/$\mathbb{S}$/$\mathbb{T}$ & \begin{tabular}[c]{@{}l@{}}A 
			node index set/sampling time-slots set of input  \\ data/ sampling  
			time-slots set of predicting  data\end{tabular} & 
			$\mathbf{M}$/$\mathcal{D}$/$\mathcal{I}$ &\begin{tabular}[c]{@{}l@{}} 
			Mask matrix/degree matrix/identity matrix \end{tabular} \\
			\midrule
			$N$/$S$/$T$ & \begin{tabular}[c]{@{}l@{}}Node size/sequence length of 
			input data/sequence \\ length  of target data\end{tabular} & $h$  & 
			Head size of multi-attention module \\
			\midrule
			$\mathbf{X}$/$\mathbf{Y}$/$\mathbf{D}$/$\mathbf{F}$ & 
			\begin{tabular}[c]{@{}l@{}} Input data/target data/historical 
			auxiliary factors/\\future auxiliary factors \end{tabular}  & $q$/$p$ 
			& Dimension of input data/auxiliary factors\\
			\bottomrule
		\end{tabular}
	}
\end{table*}

This paper focuses on performing long sequence time-series forecasting tasks on 
point traffic data. Let a node index set be \begin{small}$\mathbb{N}=\{1,2, 
\cdots, N\}$\end{small}, the sampling time-slots of input historical data at time 
\begin{small}$t$\end{small} be \begin{small}$\mathbb{S}=\{t-S+1, t-S+2, \cdots, 
t\}$\end{small}, and the sampling time-slots of predicting data be 
\begin{small}$\mathbb{T}=\{t+1, \cdots, t+T\}$\end{small}, where 
\begin{small}$S$\end{small} is the input sequence length and 
\begin{small}$T$\end{small} is predicting sequence length. The traffic data of 
node \begin{small}$i$\end{small} sampled at time \begin{small}$t$\end{small} is 
\begin{small}$X_i^{(t)}=<x_{i,1}^{(t)}, \cdots, x_{i,q}^{(t)}> \in 
R^q$\end{small}, which contains \begin{small}$q$\end{small} traffic attributes 
like traffic speed and traffic volume. Therefore, the historical traffic data can 
be represented by \begin{small}$\mathbf{X}={\{X_i^{(t)}|i\in \mathbb{N}\}}_{t\in 
\mathbb{S}}$\end{small} and future traffic data can be represented by 
\begin{small}$\mathbf{Y}={\{X_i^{(t)}|i\in \mathbb{N}\}}_{t\in 
\mathbb{T}}$\end{small}. In addition, some auxiliary attributes like sampling 
time, positional information, and weather states are absorbed to mine rich 
spatial-temporal correlation. Let the auxiliary attributes of node 
\begin{small}$i$\end{small} sampled at time \begin{small}$t$\end{small} be 
\begin{small}$D_i^{(t)}=<d_{i,1}^{(t)}, \cdots, d_{i,p}^{(t)}> \in 
R^p$\end{small}, which contains \begin{small}$p$\end{small} attributes. According 
to different sampling time, the auxiliary data can be divided into historical 
auxiliary data \begin{small}$\mathbf{D}={\{D_i^{(t)}|i\in \mathbb{N}\}}_{t\in 
\mathbb{S}}$\end{small} and future auxiliary data 
\begin{small}$\mathbf{F}={\{D_i^{(t)}|i\in \mathbb{N}\}}_{t\in 
\mathbb{T}}$\end{small}.

Therefore, we can express the traffic data forecasting tasks: \begin{small}
	$ f:\mathbf{X}\times\mathbf{D}\times\mathbf{F} \times\mathbf{G} \rightarrow 
	\mathbf{Y}$
\end{small}. In the form of a sliding window, those historical data of 
\begin{small}$S$\end{small} time-slots including traffic data 
\begin{small}$\mathbf{X}$\end{small}, historical auxiliary data 
\begin{small}$\mathbf{D}$\end{small}, future auxiliary data 
\begin{small}$\mathbf{F}$\end{small} and the graph structure 
\begin{small}$\mathbf{G}$\end{small} are fed into the forecasting model 
\begin{small}$f$\end{small} to produce the traffic states 
\begin{small}$\mathbf{Y}$\end{small} in the next \begin{small}$T$\end{small} 
time-slots.

For better understanding, frequently used notations are listed in Table 
\ref{tab:notations}.

\section{Proposed method}

As shown in Fig. \ref{fig:AIMSAN}, AIMSAN is an encoder-decoder model, mainly 
including two modules, i.e. AIM and SAN. It uses the historical traffic data 
\begin{small}$\mathbf{X}$\end{small}, historical auxiliary data 
\begin{small}$\mathbf{D}$\end{small} and future auxiliary data 
\begin{small}$\mathbf{F}$\end{small} to train its model, producing the 
predictions of future traffic data \begin{small}$\mathbf{Y}$\end{small}. 

\begin{small}
	\begin{figure*}[!htbp]
		\centering		
		\includegraphics[width=0.8\linewidth]{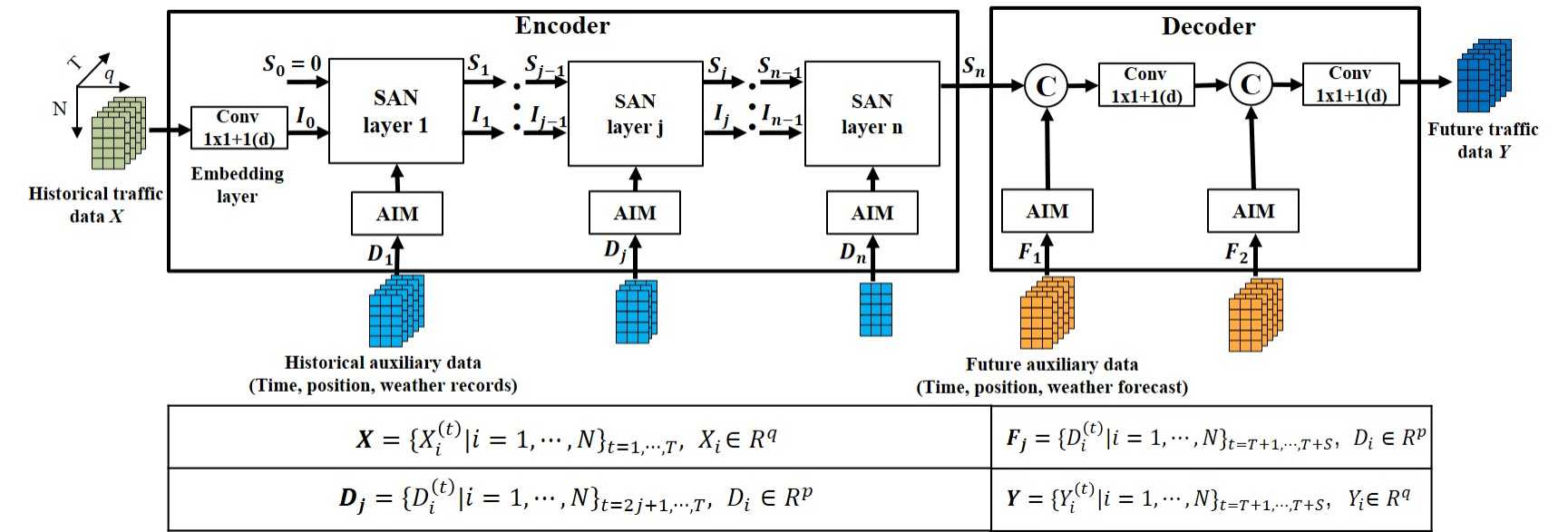}  		
		\caption{Framework of AIMSAN.
		}
		\label{fig:AIMSAN} 
	\end{figure*}
\end{small}

\subsection{Encoder structure}

Our encoder includes one embedding layer, $n$ AIMs and $n$ SAN layers. 
Specifically, historical traffic data \begin{small}$\mathbf{X}$\end{small} is fed 
into the embedding layer to increase the attribute dimension, where "Conv 
1x1+1(d)" stands for 1x1 convolutions with the dilation as 1. Due to the use of 
dilated causal convolution, the temporal dimension of hidden data in each SAN 
module decreases with each layer. The historical auxiliary data are fed into AIM 
to learn the sample and position-specific external influence. The temporal 
dimensions of historical auxiliary data are consistent with those of hidden data 
in the corresponding SAN layers. 
Then the \begin{small}${(j-1)}^{th}$\end{small} layer output data 
\begin{small}$\mathbf{I_{j-1}}$\end{small}, the feed-forward data of skip 
connection (i.e. skip data) \begin{small}$\mathbf{S_{j-1}}$\end{small} and AIM 
output \begin{small}$\mathbf{D_j}$\end{small} are fed into the 
$j^{\operatorname{th}}$ SAN layer with two outputs, 
\begin{small}$\mathbf{S_j}$\end{small} and 
\begin{small}$\mathbf{I_j}$\end{small}. 
Skip data \begin{small}$\mathbf{S_j}$\end{small} contain the hidden presentation 
of the foregoing layers, which are initiated to zeros and accumulated in each SAN 
layer. The AIM and SAN modules are elaborated next.

\subsubsection{AIM}

Traffic data is influenced by many auxiliary factors, like time (time of day, day 
of week, and holiday), position (longitude and latitude, or POI), and weather 
conditions. Therefore, AIMSAN applies AIM to analyze the influence of auxiliary 
factors. The produced weights are concatenated with hidden data in the following 
attention module by which AIMSAN can learn auxiliary information-aware attention 
values and mine more spatiotemporal dependencies of traffic data.
Fig. \ref{fig:aip} shows the structure of AIM, where the sampling time 
information, corresponding weather conditions, and node positions are fed into 
three independent embedding branches. Each branch is constructed by two 
fully-connected layers. Subsequently, the outputs of three branches are 
concatenated into one tensor \begin{small}$\mathbf{D_j}=[N, T', 3H]$\end{small} 
using matrix expansion and concatenation operations, where 
\begin{small}$N$\end{small} is the node size, \begin{small}$T'$\end{small} is the 
time-series length, and \begin{small}$H$\end{small} is the dimension of hidden 
layer. Due to the lack of POI information, positional attributes only consist of 
latitude and longitude values in this paper. Future weather information can be 
obtained from weather forecast data in practice. Thanks to the branch structure 
of AIM, we can tailor it to only mine temporal information for those open-source 
datasets without any positional information. In addition, 
it is easy to add new auxiliary factors into AIM.

\begin{small}
	\begin{figure}[!t]
		\centering
		\includegraphics[width=0.99\linewidth]{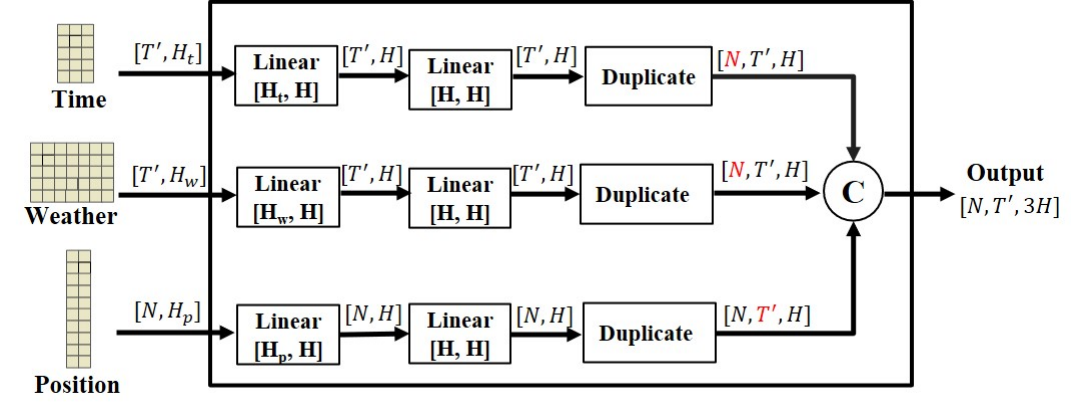}
		\caption{The structure of an AIM, where C denotes a concatenation 
		operation.}
		\label{fig:aip}       
	\end{figure}
\end{small}

\begin{small}
	\begin{figure*}[htbp]
		\centering
		\includegraphics[width=0.8\linewidth]{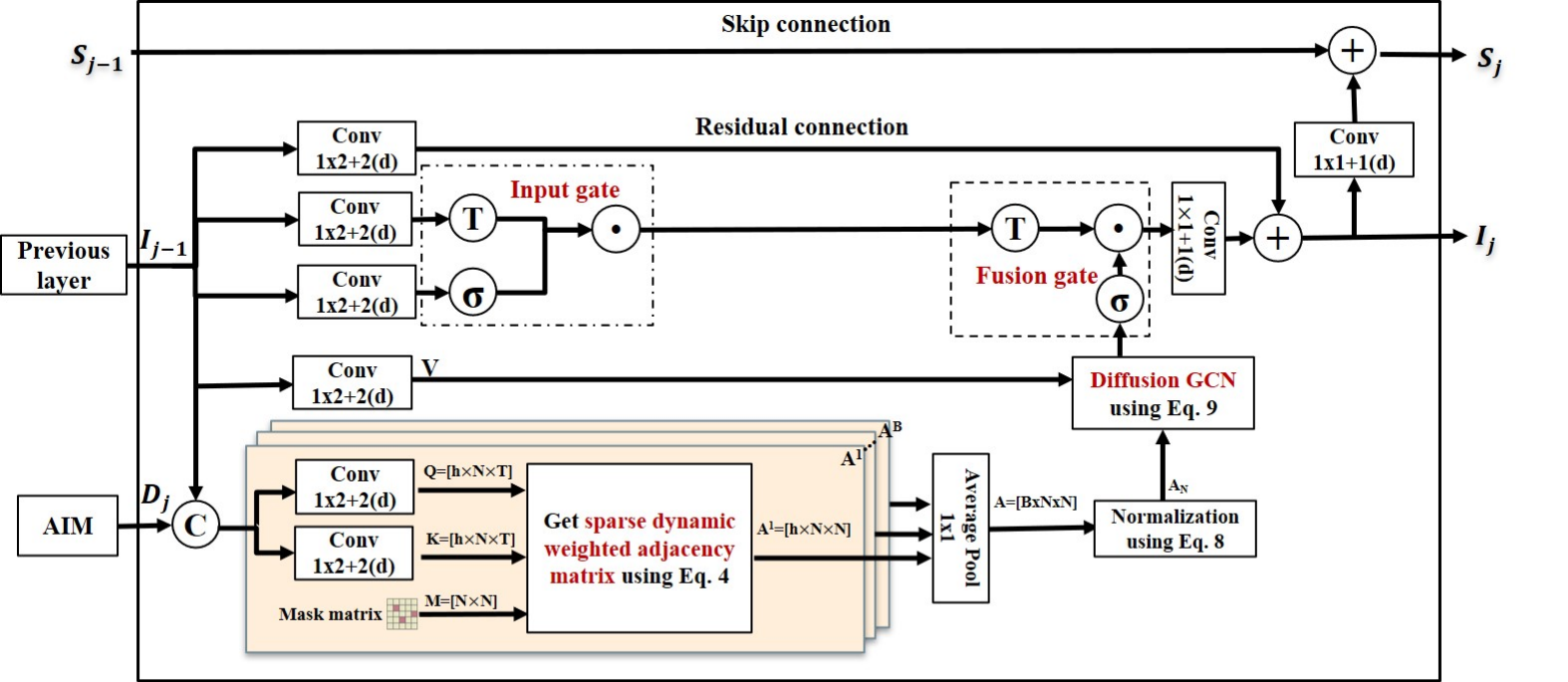}
		\caption{Frameworks of sparse cross attention-based diffusion GCN (SAN) 
		layer.}
		\label{fig:AIMSAN-module}       
	\end{figure*}
\end{small}

\subsubsection{SAN layer}
Fig. \ref{fig:AIMSAN-module} shows the framework of SAN.
It takes the previous layer output \begin{small}$\textbf{I}_{j-1}$\end{small}, 
\begin{small}$j^{\operatorname{th}}$\end{small} AIM output 
\begin{small}$\textbf{D}_{j}$\end{small} and skip data 
\begin{small}$\textbf{S}_{j-1}$\end{small} as inputs and produces the layer 
output \begin{small}$\textbf{I}_{j}$\end{small} and accumulated 
\begin{small}$\textbf{S}_{j}$\end{small}. The SAN layer mainly includes a DCC 
module, sparse attention graph convolution module and two gate structures (input 
gate and fusion gate). 

\noindent\textbf{DCC:}
First, we introduce the DCC module used in the SAN layer. 
It is used to mine temporal dependency and reduce the temporal dimension of 
hidden data, where "Conv 1x2+2(d)" stands for 1x2 convolutions with the dilation 
as 2. 
Unlike the step-by-step manner in RNN, DCC enlarges its receptive field by 
stacking the network layers, which can improve the propagation efficiency and 
avoid the potential gradient problem in RNN-based models. 

AIMSAN applies DCC in each SAN layer to learn the temporal dependency and slim 
its model size.
Taking the input temporal dimension \begin{small}$T=12$\end{small} and dilation 
factors [2, 2, 2, 2, 2, 1] for example, Fig. \ref{fig:dcc} shows the function of 
a DCC structure in the reduction dimension by ignoring the other modules in 
AIMSAN. Formally, given a 1-D sequence input \begin{small}$X\in 
	\mathbb{R}^T$\end{small} and a filter \begin{small}$f\in 
	\mathbb{R}^k$\end{small}, the dilated convolution operation of 
\begin{small}$X$\end{small} with \begin{small}$f$\end{small} at step 
\begin{small}$t$\end{small} can be represented as:

\begin{small}
	\begin{equation}
		X\ast f(t)=\sum_{i=0}^{k-1}f(i)\cdot X(t-d \times i),
		\label{eq:conv}
	\end{equation} 
\end{small} 

\noindent where $\ast$ denotes the convolution operation, 
\begin{small}$d$\end{small} represents the dilation factor, 
\begin{small}$k$\end{small} represents the size of the filter, and the numbers in 
parenthesis (\begin{small}$t-d\times i$\end{small}) indicate the indices of the 
vectors. When \begin{small}$d=1$\end{small}, dilation convolution is equivalent 
to common causal convolution. 
As shown in Fig. \ref{fig:dcc}, the temporal dimensions of those feed-forward 
data are reduced gradually after each DCC, and finally converge to 1. Therefore, 
the feed-forward data in the last SAN layer contain all the temporal information 
in the past \begin{small}$T$\end{small} time slots.

\begin{small}
	\begin{figure}[htbp]
		\centering
		\includegraphics[width=1\linewidth]{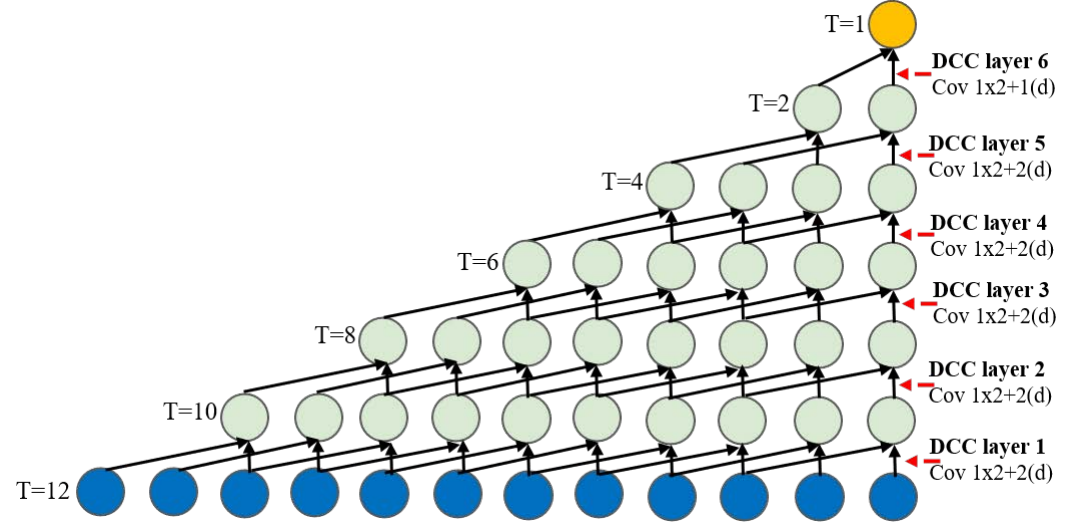}  
		\caption{Framework of dilated causal convolution structure, taking the 
		input temporal dimension \begin{small}$T$\end{small}=12 and dilation 
		factors as [2, 2, 2, 2, 2, 1] as an example.
		}
		\label{fig:dcc} 
	\end{figure}
\end{small}

\noindent\textbf{GCN operations:}
Inspired by the weighted matrix in the cross-attention module of Transformer 
\cite{NIPS2017_3f5ee243}, AIMSAN calculates the auxiliary-information-aware 
weighted matrix as the adaptive adjacent matrix for the graph convolution 
operation.

\begin{equation}
	\mathbf{I}_{j}' = \mathbf{I}_j \; || \; \mathbf{D}_{j+1}
	\label{eq:cat}
\end{equation}
\begin{equation}
	\mathbf{A}^b = FF(\mathbf{I}_j')\times FF(\mathbf{I}_{j}')^T, \; \mathbf{A}^b 
	\in \mathbb{R}^{N\times N}
	\label{eq:A_head}
\end{equation}

In (\ref{eq:cat}), AIMSAN concatenates the output of the 
\begin{small}$j^{th}$\end{small} SAN layer 
\begin{small}$\textbf{I}_{j}$\end{small} with historical embedded auxiliary 
information \begin{small}$\textbf{D}_{j+1}$\end{small} in node size dimension, 
where \begin{small}$||$\end{small} is vector concatenation operator, and  
\begin{small}$\textbf{I}_{j}'$\end{small} is the results. 
The calculation process of the attention-based weighted matrix is shown in 
(\ref{eq:A_head}), where the FF() is the feed-forward function (the convolution 
operation is used here). After $h$ times multiplication, AIMSAN obtains the 
\begin{small}$h$\end{small}-head attention-based weighted tensor 
\begin{small}$\textbf{A}^b = [\textbf{A}_1,\cdots,\textbf{A}_h]$\end{small} of 
size \begin{small}$h \times N \times N$\end{small} for each mini-batch data.

However, the common attention module suffers quadratic computation complexity, 
thus influencing the scalability of the algorithm toward large-scale traffic 
tasks. Therefore, considering the sparse correlation of traffic nodes, AIMSAN 
applies a sparse strategy to the computing process of weighted matrices, i.e., 

\begin{small}
	\begin{equation}
		\mathbf{A}^b = f_s(\mathbf{Q},\mathbf{K},\mathbf{M})
		\label{eq:atten_mx1}
	\end{equation}
\end{small}
\begin{small}
	\begin{equation}
		\mathbf{Q}=(w_{i,j}^1)_{N \times T},\; \mathbf{K}=(w_{i,j}^2)_{N \times 
		T},\;
		\mathbf{M}=(m_{i,j})_{N \times N}
		\label{eq:atten_mx2}
	\end{equation}
\end{small}

\begin{small}
	\begin{equation}
		a_{ij}^b= m_{i,j} \times 
		\sum_{x=1}^{T}{w_{i,x}^1{{(w}_{j,x}^2)}^T},
		\label{eq:atten_mx3}
	\end{equation}
\end{small}

\begin{small}
	\begin{equation}
		m_{i,j} \in \mathbf{M}, \;
		w_{i,x}^1 \in \mathbf{Q}, \;
		w_{j,x}^2 \in \mathbf{K}, \;
		a_{ij}^b \in \mathbf{A}^b,\;
		b \in [1,\cdots,B]
		\label{eq:atten_mx4}
	\end{equation}
\end{small}

\noindent where
\begin{small}$f_s()$\end{small} is the sparse strategy function, and its 
calculation procedure is shown in (\ref{eq:atten_mx2}-\ref{eq:atten_mx4}). 
\begin{small}$\mathbf{Q}$\end{small} and \begin{small}$\mathbf{K}$\end{small} are 
the output of two feed-forward functions in (\ref{eq:A_head}). 
\begin{small}$\mathbf{M}$\end{small} is the mask adjacent matrix. For simplicity, 
we describe the computing process in the head dimension (i.e., the case of one 
head attention \begin{small}$h=1$\end{small} in Fig. \ref{fig:AIMSAN-module}).

Subsequently, AIMSAN performs an Average Pooling operation on all 
\begin{small}$\textbf{A}^b$\end{small} on the head dimension and then obtains the 
dynamic adjacent matrix 
\begin{small}$\textbf{A}={[\textbf{A}^{1'},\cdots,\textbf{A}^{B'}]}_{B\times N 
\times N}$\end{small} of the current layer.

After finding the adaptive weights of adjacent matrix 
\begin{small}$\mathbf{A}$\end{small}, AIMSAN obtains a normalized graph Laplacian 
of \begin{small}$\mathbf{A}$\end{small} (\begin{small}$\mathbf{A}_N$\end{small}) 
by using 

\begin{equation}
	\mathbf{A}_N = \mathcal{D}^{\frac{1}{2}} \times (\mathbf{A}+\mathcal{I}) 
	\times \mathcal{D}^{\frac{-1}{2}},
	\label{eq:norm}
\end{equation}

\noindent where \begin{small}$\mathcal{I}$\end{small} is the identity matrix, and 
\begin{small}$\mathcal{D}$\end{small} denotes the degree matrix of 
\begin{small}$\mathcal{A+I}$\end{small}.

In many existing studies, diffusion convolutional operation plays an important 
role in spatial-temporal modeling \cite{ijcai2019-264,li2018diffusion}. 
Therefore, AIMSAN also applies diffusion graph convolutions to extract more 
spatial information. 
The computing process of diffusion graph convolution is

\begin{small}
	\begin{equation}
		\mathbf{Z} = \sum_{i=0}^{k} \mathbf{A}_N^i \mathbf{X} \mathbf{W}_i,
		\label{eq:2order}
	\end{equation}
\end{small}

\noindent where \begin{small}
	$\mathbf{A}_N$\end{small} is the normalized adjacent matrix, 
	\begin{small}$k$\end{small} is the diffusion step, \begin{small}
	$\mathbf{X}$\end{small} is the input data, and 
	\begin{small}$\mathbf{W}_i$\end{small} is the learnable weights.

\noindent\textbf{Gate structures:}
Two gate structures, input gate and fusion gate, are used to improve the 
generalization ability of AIMSAN. The input gate selectively passes feed-forward 
data, and the fusion gate fuses the information extracted by the GCN module and 
input gate. Formally, the gate unit can be represented as: 

\begin{small}
	\begin{equation}
		O = T(X_1) \odot \sigma(X_2),  
		\label{eq:gates}
	\end{equation}
\end{small}

\noindent where $\odot$ denotes the Hadamard product, 
T(\begin{small}$\cdot$\end{small}) represents an activation function and TanH 
\cite{shakiba2020novel} is used in the following experiments. \begin{small}
	$\sigma(\cdot)$  
\end{small} means the Sigmoid function that is used to select the information 
passing through the layer, and \begin{small}$X_1$\end{small} and 
\begin{small}$X_2$\end{small} are two input data of gating structure generated 
from different feed-forward layers.


\subsection{Decoder structure}
Our encoder structure abstracts the high-level representation of the historical 
traffic data and auxiliary data after stacking multiple layers and encodes them 
into one tensor (i.e. the last skip data \begin{small}$\mathbf{S_n}$\end{small}). 
Then \begin{small}$\mathbf{S_n}$\end{small} is fed into our decoder structure. As 
shown in Fig. \ref{fig:AIMSAN}, the decoder consists of two output layers and two 
AIMs. The future auxiliary data, including sample-specific time, position and 
future weather information, are fed into each AIM to increase the awareness of 
future states for AIMSAN. 
The output of AIM is concatenated with the feed-forward data and then 
sequentially fed into each output layer. Using the "Conv 1x1+1(d)" operation, 
output layers reduce the dimension of feed-forward data to fit the target data.

\subsection{Complexity Analysis}

Calculation of dynamic neighbor matrix of size \begin{small}$N \times 
N$\end{small} is the dominant factor leading to high computation complexity, 
where \begin{small}$N$\end{small} is the node size. 
Fig. \ref{fig:nxn} shows examples of the normal adjacency matrix and sparse 
adjacency one, where the node size is 5. In Fig. \ref{fig:nxn1}, the weights 
between each pair of nodes are calculated. While in the sparse adjacency matrix, 
each node only calculates its weights with the top-$k$ related neighbors, as 
shown in Fig. \ref{fig:nxn2} ($k$=3).   

\begin{small}
	\begin{figure}[htbp]
		\centering
		\subfigure[Normal adjacency matrix]{
			\includegraphics[width=0.45\linewidth]{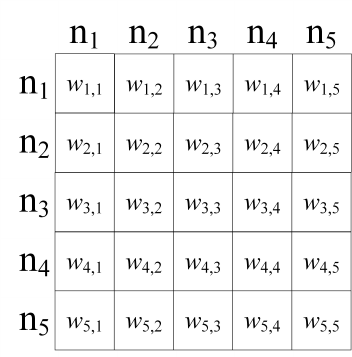}
			\label{fig:nxn1}
		}
		\subfigure[Sparse adjacency matrix]{
			\includegraphics[width=0.45\linewidth]{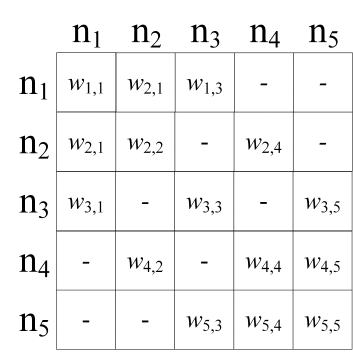}
			\label{fig:nxn2}
		}
		\caption{Examples of the normal adjacency matrix and sparse adjacency 
		matrix, where node size is 5, and the number of most related nodes is 3 
		in (b).
		}
		\label{fig:nxn} 
	\end{figure}
\end{small}

\noindent\textbf{Computational complexity of existing models}:
Some dynamic GCN methods use the matrix manipulation of two embedding matrices 
\begin{small}$\mathbf{E_1}$\end{small} and \begin{small}$\mathbf{E_2}$\end{small} 
to obtain a dynamic adjacent matrix \begin{small}$\mathbf{A}$\end{small} as shown 
in Fig. \ref{fig:nxn1}, 

\begin{small}
	\begin{align}
		\mathbf{A}=\mathbf{E_1}&\mathbf{E_2}^T 
		\label{eq:dynamic}\\ \mathbf{E_1},\mathbf{E_2}\in \mathbb{R}^{N\times H}, 
		\; & \mathbf{A} \in \mathbb{R}^{N\times N}
		\label{eq:dynamic2}
	\end{align}
\end{small}

\noindent where \begin{small}$H$\end{small} is the hidden data size 
\cite{DMSTGCN}. Some methods apply a learnable matrix, which ignore the spatial 
sparsity of nodes and require a quadratic computational complexity in a graph 
convolution process \cite{li2021spatial}. In addition, attention-based algorithms 
use the data-driven method to construct a weighted matrix dynamically 
\cite{lu2022make},
i.e., 

\begin{small}
	\begin{align}
		Attention(\mathbf{Q}, \mathbf{K}, 
		\mathbf{V})&=\text{softmax}(\frac{\mathbf{Q}\mathbf{K}^T}{\sqrt{d_k}})\mathbf{V}.
		\label{eq:transformer}\\
		\mathbf{Q}, \mathbf{K}, \mathbf{V}&\in \mathbb{R}^{N\times H}
		\label{eq:transformer2}
	\end{align}
\end{small}

\begin{small}$\mathbf{Q}$\end{small}, \begin{small}$\mathbf{K}$\end{small}, and 
\begin{small}$\mathbf{V}$\end{small} are Queries, Keys, and Values in the 
attention modules, respectively. From (\ref{eq:transformer}), 
\begin{small}$\mathbf{Q}\mathbf{K}^T$\end{small} calculates the scores of each 
pair of nodes, producing a weight matrix like the one in Fig. \ref{fig:nxn1}. 
Therefore, those attention-based methods also have 
\begin{small}$\mathcal{O}(N^2)$\end{small} computational complexity in 
calculating the attention values.

\noindent\textbf{Computational complexity of AIMSAN }:
The proposed SAN layer applies the spatial sparsity of nodes into the 
cross-attention process, by using (\ref{eq:atten_mx3}). Here, we analyse the 
computational complexity in calculating \begin{small}$\mathbf{A^i}$\end{small} in 
(\ref{eq:atten_mx1}).
Let \begin{small}$\mathbf{Q}=(w_{i,j}^1)_{N \times T}$\end{small}, 
\begin{small}$\mathbf{K}=(w_{i,j}^2)_{N \times T}$\end{small}
be the outputs of two feed-forward function in (\ref{eq:A_head}), and the mask 
matrix be \begin{small}$\mathbf{M}=(m_{i,j})_{N \times N}$\end{small}.
The mask matrix can be obtained from a distance-based adjacent matrix where only 
connected nodes have weights or from other context-based ones that only retain 
the weights from the top-\begin{small}$k$\end{small} most related neighbors for 
each node.
Therefore, for each traffic node, we only need to calculate the weight values for 
its \begin{small}$k$\end{small} related nodes and obtain the sparse adjacent 
matrix as shown in Fig. \ref{fig:nxn2}, which exhibits 
\begin{small}$\mathcal{O}(N\times k)$\end{small} computational complexity, and 
\begin{small}$k\ll N$\end{small} in practice.

\section{Experiments}
This section first introduces the data sets, baselines, and evaluation metrics 
used in this paper. Then, we compare the performance of AIMSAN on three traffic 
datasets with its competitive peers. Finally, we perform the parameter 
sensitivity analysis and ablation experiments for AIMSAN. 

\subsection{Dataset Description}
We verify the performances of AIMSAN on three real-world traffic datasets 
(METR-LA, PEMS04, and PEMS07) with different node sizes. We list the details of 
three datasets in Table \ref{tab:dataset}.  

\begin{table*}[htbp]\footnotesize
	\caption{Summary statistics of datasets}
	\label{tab:dataset}
	\resizebox{\linewidth}{!}{%
		\begin{tabular}{ccccccc}
			\toprule
			Datasets & Date   & \#Nodes & \#Time Steps &Used attribute &Target 
			attribute  & Period \\  
			\toprule
			METR-LA  & 2012/03/01-2012/05/31 & 207     & 34272   &Speed, time, 
			position, weather   &Speed          & 5mins  \\
			PEMS04   & 2018/01/01-2018/02/28 & 307     & 16992   &Volume,  time   
			&Volume  & 5mins  \\
			PEMS07   & 2012/05/01-2012/08/07 & 883     & 28224   &Volume,  time   
			&Volume  & 5mins  \\
			\toprule
		\end{tabular}%
	}
\end{table*}

METR-LA contains traffic speed information collected from 207 loop detectors on 
the highway of Los Angeles County, whose period is March 1st to June 30th, 2012 
\cite{jagadish2014big}. 
PEMS04 and PEMS07 are sampled from Performance Measurement System (PEMS) 
\footnote{https://pems.dot.ca.gov/} on the freeway system spanning all major 
metropolitan areas of California. PEMS04 uses 307 sensors in the San Francisco 
Bay Area, and its frequently referenced period is January 1st to February 28th, 
2018. PEMS07 contains 883 sensors in the Los Angeles Area, and its frequently 
referenced period is May to June 2012.

In the data preparation, we enrich the weather attribute of METR-LA according to 
the available node location information (longitude and latitude). So, we use 
traffic speed, time (time of day, day of week, and holiday), position, and 
weather information (temperature, humidity, and weather condition) for METR-LA, 
and set the speed as the target attribute. Due to the lack of position 
information for PEMS04 and PEMS07, we cannot obtain their weather information. 
So, AIMSAN only uses traffic volume and time information (time of day, day of 
week, and holiday) to predict future traffic volume. As shown in Fig. 
\ref{fig:aip}, AIM can easily adjust to cases of lacking different auxiliary 
information. Besides, three datasets are divided into the training set (70\%), 
validation set (10\%), and test set (20\%) in chronological order.

\subsection{Baselines and Evaluation metrics}
We compare AIMSAN with the following models.
\begin{itemize}
	\item Graph WaveNet \cite{ijcai2019-264}: Graph WaveNet is a convolution 
	network architecture, which introduces a self-adaptive graph to capture the 
	hidden spatial dependency, and uses dilated convolution to capture the 
	temporal dependency.
	\item DMSTGCN \cite{DMSTGCN}: It is a dynamic graph neural network model that 
	considers the influence traffic volumes on traffic speed prediction. 
	Therefore, It designs a dual-branch structure for learning two different 
	factors and uses a fusion strategy to obtain the multi-faceted 
	spatial-temporal characteristics inherent in traffic data. In METR-LA, there 
	is only one attribute (traffic speed). Therefore, temporal information is 
	used.
	
	\item ST-MetaNet \cite{9096591}:  It is a deep meta-learning-based model 
	containing the meta graph attention network and meta RNN. The weights of the 
	two networks are dynamically generated from the embeddings of geo-graph 
	attributes and the traffic context learned from dynamic traffic states. While 
	PEMS04 and PEMS07 have no positional information, only other geo-graph 
	attributes can be used.  
	
	\item TGANet \cite{comcom}: It is our previous work, which combines dilated 
	temporal convolution, multi-kernel diffusion graph convolution, and sparse 
	self-attention module to learn the spatial-temporal dynamics of traffic data 
	from different perspectives.
\end{itemize}

Three standard metrics in traffic forecasting are 1) Mean Absolute Error (MAE), 
2) Root Mean Squared Error (RMSE), and 3) Mean Absolute Percentage Error (MAPE).

\begin{small}
	\begin{equation}
		MAE = \frac{1}{n} \sum\limits_{i=1}^n| \hat{y}_i-y_i|
	\end{equation} 
	\begin{equation}
		RMSE = \sqrt{\frac{1}{n} \sum\limits_{i=1}^n| \hat{y}_i-y_i |^2}
	\end{equation} 
	\begin{equation}
		MAPE= \frac{100\%}{n} \sum\limits_{i=1}^n| \frac{\hat{y}_i-y_i}{y_i}|
	\end{equation}
\end{small}

\noindent{where} \begin{small}$y_i$\end{small} and 
\begin{small}$\hat{y}_i$\end{small} are the real and predicted values at the 
\begin{small}$i^{th}$\end{small} time slot, respectively. 
\begin{small}$n$\end{small} is the number of time slots. Specifically,  the 
values of MAE, RMSE, and MAPE are used to measure the prediction errors. The 
smaller, the better. 

Besides, we propose the Improvement score (\begin{small}$\Gamma$\end{small}) to 
compare two methods:

\begin{equation}
	\Gamma = \frac{m_1-m_2}{m_1}\times 100\%
	\label{eq:impro}
\end{equation}

\noindent where \begin{small}$m_1$\end{small} and \begin{small}$m_2$\end{small} 
are metric values (like MAE) of two methods (\begin{small}$M_1$\end{small} and 
\begin{small}$M_2$\end{small}). The positive value of 
\begin{small}$\Gamma$\end{small} denotes that \begin{small}$M_2$\end{small} is 
better than \begin{small}$M_1$\end{small}, otherwise not.

\subsection{Hyper-parameters and experimental settings}
There are two important hyper-parameters in AIMSAN, the head size of the 
attention module (\begin{small}$h=3$\end{small}), and the output dimension of AIM 
(\begin{small}$d_{\operatorname{AIM}}=16$\end{small}). The batch size is set as 
32. We apply the Adam optimizer to train AIMSAN with 100 epochs. The initial 
learning rate is set to 1e-3 and decays to one-tenth per twenty epochs with a 
minimum of 1e-6. The dilated factors of TCN are set to [2, 2, 2, 2, 2, 1]. The 
hidden dimension of the feed-forward network is 32, and the hidden dimension of 
the skip connection is 256. We set the diffusion graph convolution order as 
\begin{small}$k= 2$\end{small}. After experimental comparisons, we use TanH and 
Sigmoid functions in the Fusion gate and set the hidden size of AIM as 16 (see 
more details in Appendix A in our supplementary file 
All the experiments are conducted on a GPU server with a single PNY GeForce RTX 
2080 Ti (12GB memory) GPU.

\subsection{Comparative experiments}

\begin{small}
	\begin{table*}[t] \footnotesize
		\caption{Performance comparison of baseline models and AIMSAN. The best 
		results are highlighted in \textbf{bold}, and the second-best results are 
		\underline{underlined}.}
		\label{tab:comp}
		\resizebox{\textwidth}{!}{%
			\begin{tabular}{cccccccccccccc}
				\toprule
				\multirow{2}{*}{Data} & \multirow{2}{*}{Model} & 
				\multicolumn{4}{c}{MAE} & \multicolumn{4}{c}{RMSE} & 
				\multicolumn{4}{c}{MAPE(\%)} \\
				\cmidrule(r){3-6} \cmidrule(r){7-10} \cmidrule(r){11-14}&    & 
				15min & 30min & 60min & all   & 15min & 30min & 60min & all & 
				15min & 30min & 60min & all   \\
				\midrule
				\multirow{5}{*}{METR-LA} 
				& GraphWaveNet (2019) & \underline{2.70} & \underline{3.10} & 
				3.56 & 3.06 & \underline{5.18} & \underline{6.24} & 7.37 & 
				\underline{6.11} & \textbf{6.93} & \underline{8.43} & 10.07 & 
				\textbf{8.27} \\
				& DMSTGCN (2021) & 2.84 & 3.24 & 3.68 & 3.18 & 5.52 & 6.51 & 7.48 
				& 6.34 & 7.50 & 9.06 & 10.72 & 8.86 \\
				& TGANet (2022) & 2.74 & \underline{3.10} & \underline{3.49} & 
				\underline{3.05} & 5.30 & 6.31 & \underline{7.31} & 6.16 & 7.09 & 
				8.44 & \textbf{9.99} & 8.31 \\
				& ST-MetaNet (2022) &\textbf{2.69} & \underline{3.10} & 3.58 & 
				3.06 & \textbf{5.15} & 6.27 & 7.52 & 6.14 & \underline{6.94} & 
				8.64 & 10.82 & 8.56 \\
				& AIMSAN (ours) & 2.73 & \textbf{3.08} & \textbf{3.47} & 
				\textbf{3.04} & 5.23 & \textbf{6.20} & \textbf{7.19} & 
				\textbf{6.06} & 6.99 & \textbf{8.41} & \underline{10.02} & 
				\underline{8.28} \\
				\midrule
				\multirow{5}{*}{PEMS04} 
				& Graph Wavenet (2019) & 18.77 & 20.21 & 22.67 & 20.26 & 29.88 & 
				31.89 & 35.18 & 31.92 & 13.06 & 14.30 & 15.94 & 14.16 \\
				& DMSTGCN (2021) & \textbf{17.88} & \textbf{18.71} & 20.49 & 
				\textbf{18.77} & \textbf{28.76} & \textbf{30.07} & \textbf{32.53} 
				& \textbf{30.09} & \textbf{12.28} & \textbf{12.93} & 14.36 & 
				\textbf{12.96} \\
				& TGANet (2022) & 18.76 & 19.42 & 20.52 & 19.34 & 29.83 & 
				\underline{31.09} & 32.76 & 30.91 & 12.84 & 13.59 & 
				\underline{14.18} & 13.34 \\
				& ST-MetaNet (2022) &19.06 & 20.36 & \underline{22.46} & 20.31 & 
				30.13 & 32.05 & 34.98 & 31.95 & 13.10 & 13.98 & 15.48 & 13.98 \\
				& AIMSAN (ours) & \underline{18.66} & \underline{19.32} & 
				\textbf{20.30} & \underline{19.24} & \underline{29.81} & 31.10 & 
				\underline{32.54} & \underline{30.88} & \underline{12.82} & 
				\underline{13.20} & \textbf{13.87} & \underline{13.16} \\
				\midrule
				\multirow{5}{*}{PEMS07}
				& Graph Wavenet (2019) & \underline{19.81} & 21.99 & 25.68 & 
				22.05 & \underline{31.82} & 35.15 & 40.25 & 35.06 & 
				\underline{8.45} & 9.40 & 11.11 & 9.46 \\
				& DMSTGCN (2021) & - & - & - & - & - & - & - & - & - & - & - & - 
				\\
				& TGANet (2022) & 20.29 & \underline{21.52} & \underline{23.51} & 
				\underline{21.41} & 32.32 & \underline{34.77} & \underline{38.34} 
				& \underline{34.58} & 8.75 & \underline{9.21} & \underline{10.05} 
				& \underline{9.19} \\
				& ST-MetaNet (2022) & - & - & - & - & - & - & - & - & - & - & - & 
				- \\
				& AIMSAN (ours) & \textbf{19.79} & \textbf{20.99} & 
				\textbf{22.82} & \textbf{20.90} & \textbf{31.98} & \textbf{34.54} 
				& \textbf{37.99} & \textbf{34.33} & \underline{\textbf{8.36}} & 
				\textbf{8.86} & \textbf{9.74} & \textbf{8.85} \\
				\midrule
			\end{tabular}%
		}
	\end{table*}
\end{small}

In this subsection, we compare the proposed method with four state-of-art 
GCN-based peers in terms of prediction accuracy, memory usage and runtime cost. 
In addition, we compare AIMSAN with four classic models. Due to the poor 
performance of those classic models, we present the related results in Appendix B 
in our supplementary file.

\subsubsection{Prediction accuracy}

Table \ref{tab:comp} shows the experimental results of AIMSAN and four 
state-of-art GCN models on the aforementioned three traffic datasets. The target 
attribute of METR-LA is traffic speed, but that of PEMS04 and PEMS07 is traffic 
volume. Each experiment is repeated five times. The average values of MAE, RMSE, 
and MAPE at the 3rd, 6th, 12th steps, and all steps are recorded in Table 
\ref{tab:comp}, respectively.

Graph Wavenet and TGANet construct multiple weighted adjacent matrices to mine 
rich spatial-temporal correlation of traffic data. Specifically, Graph WaveNet 
applies fixed adjacent matrices (predefined or learned) once the training is 
finished. TGANet and AIMSAN construct adaptive adjacent matrices by using 
attention strategies. However, AIMSAN considers the dynamic of traffic data under 
different cases of auxiliary factors (time, position, or weather). Therefore, it 
constructs AIM to embed auxiliary information of the current time window and 
learns dynamic adjacent matrix combining the above embedding information. As a 
result, AIMSAN outperforms the other two methods on the three datasets.

Both AIMSAN and DMSTGCN use auxiliary information to improve prediction accuracy. 
The former uses auxiliary factors like time and weather, while the latter 
combines secondary traffic attribute. Specifically, DMSTGCN builds two branches 
to learn the hidden correlations between the target and secondary traffic 
attributes. 
In METR-LA, there is only a traffic speed attribute without any secondary one. 
Therefore the performance of DMSTGCN is worse than other deep-learning methods, 
which reveals that DMSTGCN is sensitive to secondary attribute. In PEMS04 and 
PEMS07, traffic speed and volume are available. Therefore, in PEMS04, DMSTGCN 
outperforms others in most cases, revealing that different traffic parameters 
contain hidden correlations and can be used for target attribute prediction. 
\textit{It's worth noting that AIMSAN and other deep learning methods only use 
one kind of traffic attribute, which fails to mine the correlation among 
different traffic attributes.} When comparing those methods using one kind of 
traffic attribute on PEMS04, AIMSAN has the best performance because it learns 
the hidden feature of auxiliary factors of traffic data and constructs an 
auxiliary-information-aware adjacent matrix for dynamic GCN operations.
In addition, for DMSTGCN, an out-of-memory error occurs in PEMS07 due to its 
dual-branch structure and the large-scale node size of PEMS07. In AIMSAN, we use 
a sparse strategy to make it scalable toward the different scales of traffic 
tasks. In conclusion, 
when dealing with traffic prediction tasks under different node scales and data 
availability, AIMSAN outperforms other baselines in most cases.

\begin{small}
	\begin{table*}[!htbp] \footnotesize
		\centering
		\caption{Comparison of AIMSAN, TGANet and DMSTGCN on overall performance. 
		The best results are highlighted in \textbf{bold}, and the second-best 
		results are \underline{underlined}.}
		\label{tab:overall_performance}
		\begin{tabular}{cccccccccc}
			\toprule
			\multirow{2}{*}{Method} & \multicolumn{3}{c}{METR-LA 
			(\begin{small}$N$\end{small}=207)} & \multicolumn{3}{c}{PEMS04 
			(\begin{small}$N$\end{small}=307)} & \multicolumn{3}{c}{PEMS07 
			(\begin{small}$N$\end{small}=883)} \\
			\cmidrule(r){2-4} \cmidrule(r){5-7} \cmidrule(r){8-10} & MAE & MAPE & 
			RMSE & MAE & MAPE & RMSE & MAE & MAPE & RMSE \\
			\midrule
			AIMSAN (ours) & \textbf{3.04} & \textbf{6.06} & \textbf{8.28} & 
			\underline{19.24} & \underline{30.88} & \underline{13.16} & 
			\textbf{20.90} & \textbf{34.33} & \textbf{8.85} \\
			\midrule
			DMSTGCN (2021) & 3.18 & 6.34 & 8.86 & \textbf{18.77} & \textbf{30.09} 
			& \textbf{12.96} & - & - & - \\
			$\Gamma$ & 4.40\% & 4.42\% & 6.55\% & -2.50\% & -2.63\% & -1.54\% & - 
			& - & - \\
			\midrule
			TGANet (2022) & \underline{3.05} & 6.16 & \underline{8.31} & 19.34 & 
			30.91 & 13.34 & - & - & - \\
			$\Gamma$  & 0.33\% & 1.62\% & 0.36\% & 0.52\%  & 0.10\%  & 1.35\% & - 
			& - & - \\
			\midrule
			ST-MetaNet (2022) & 3.06  & \underline{6.14}  & 8.56 & 20.31  & 
			31.95  & 13.98 & - & - & - \\
			$\Gamma$  & 0.65\% & 1.30\% & 3.27\% & 5.27\%  & 3.35\%  & 5.87\%  & 
			- & - & - \\
			\midrule
		\end{tabular}
	\end{table*}
\end{small}

\begin{small}
	\begin{figure*}[t] \footnotesize
		\centering
		\subfigure[Maximum GPU memory occupancy]{
			\includegraphics[width=0.31\linewidth]{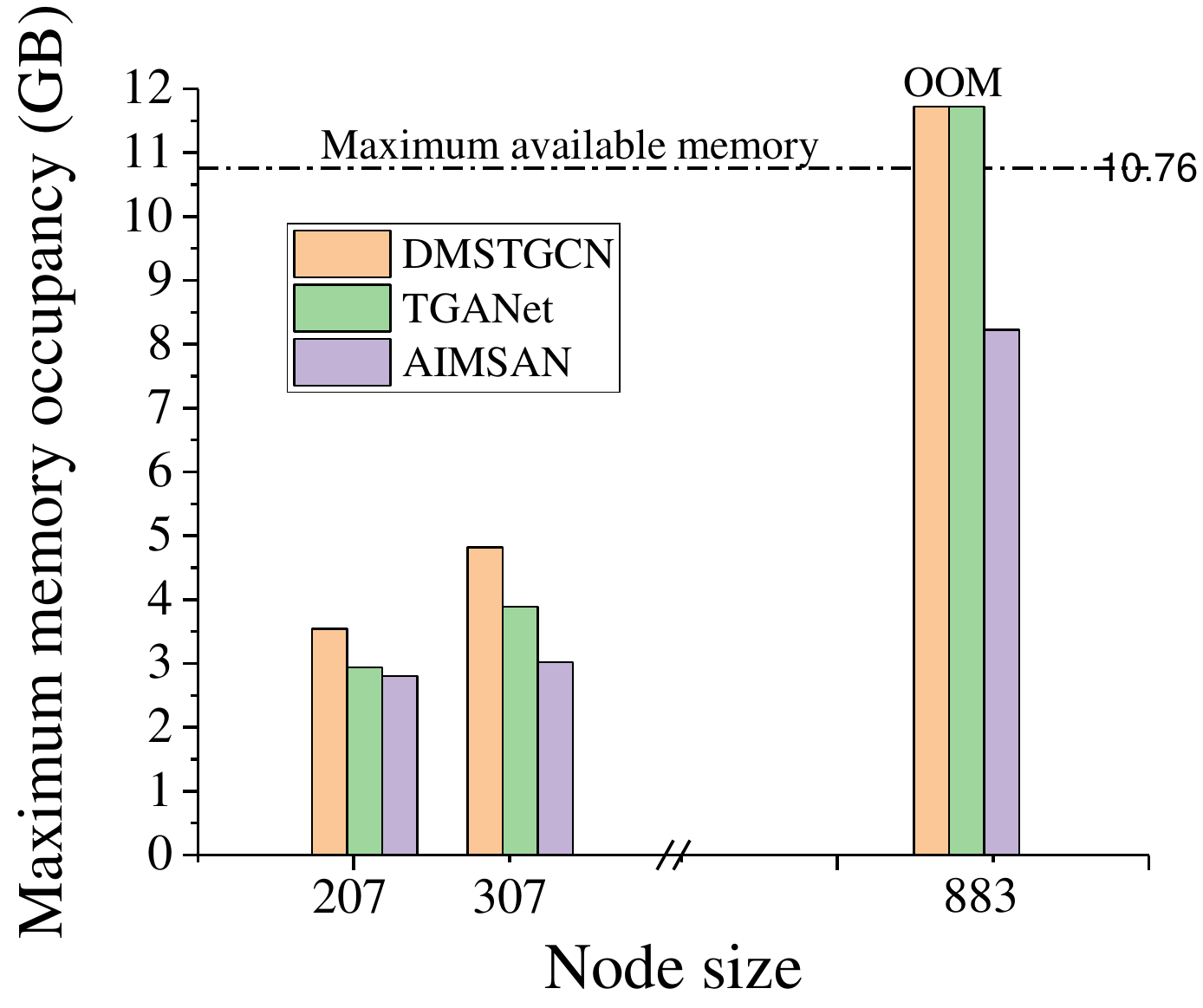}
			\label{fig:meo}
		}
		\centering
		\subfigure[Training time per batch]{
			\includegraphics[width=0.31\linewidth]{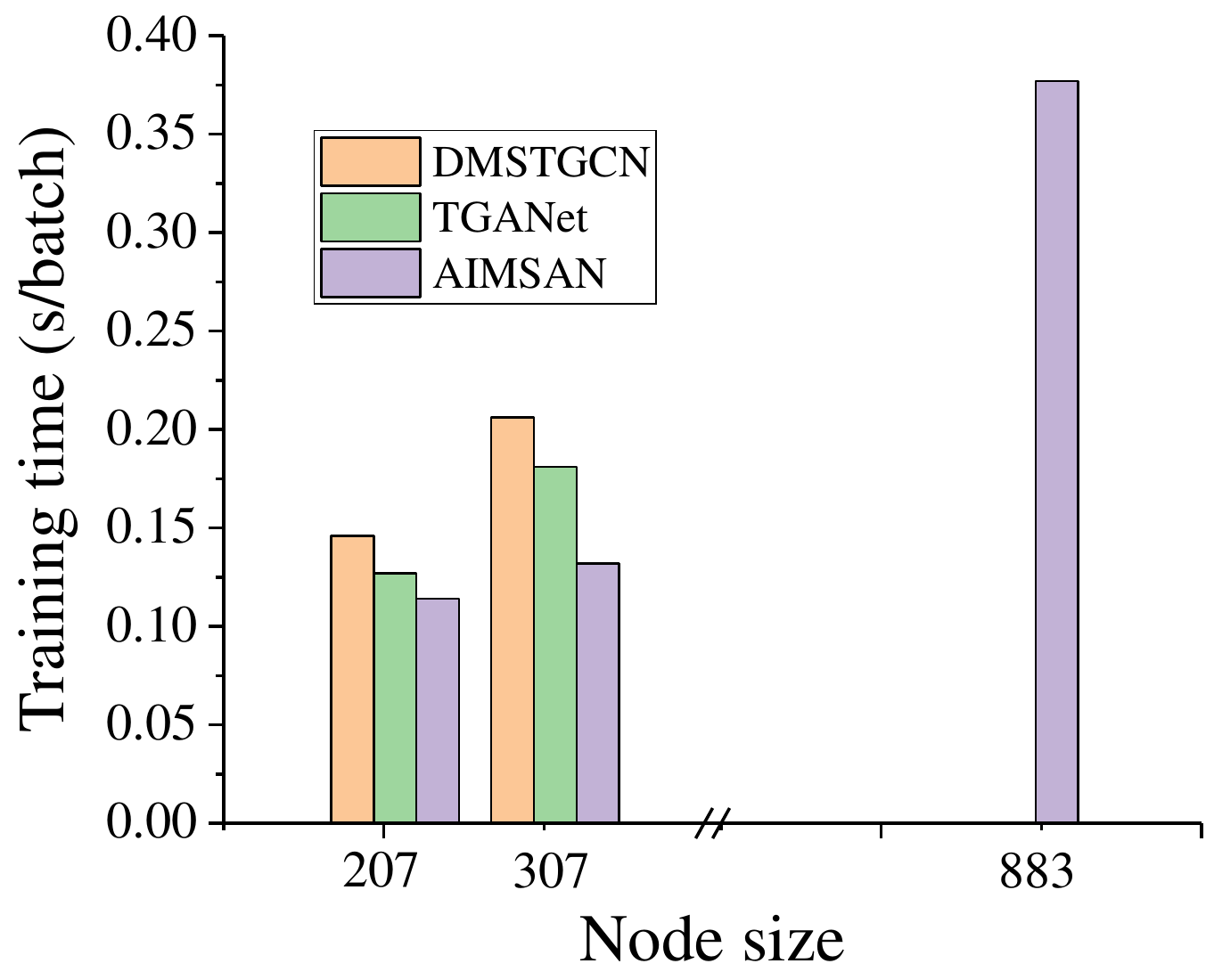}
			\label{fig:traintime}
		}
		\centering
		\subfigure[Validating time per batch]{
			\includegraphics[width=0.31\linewidth]{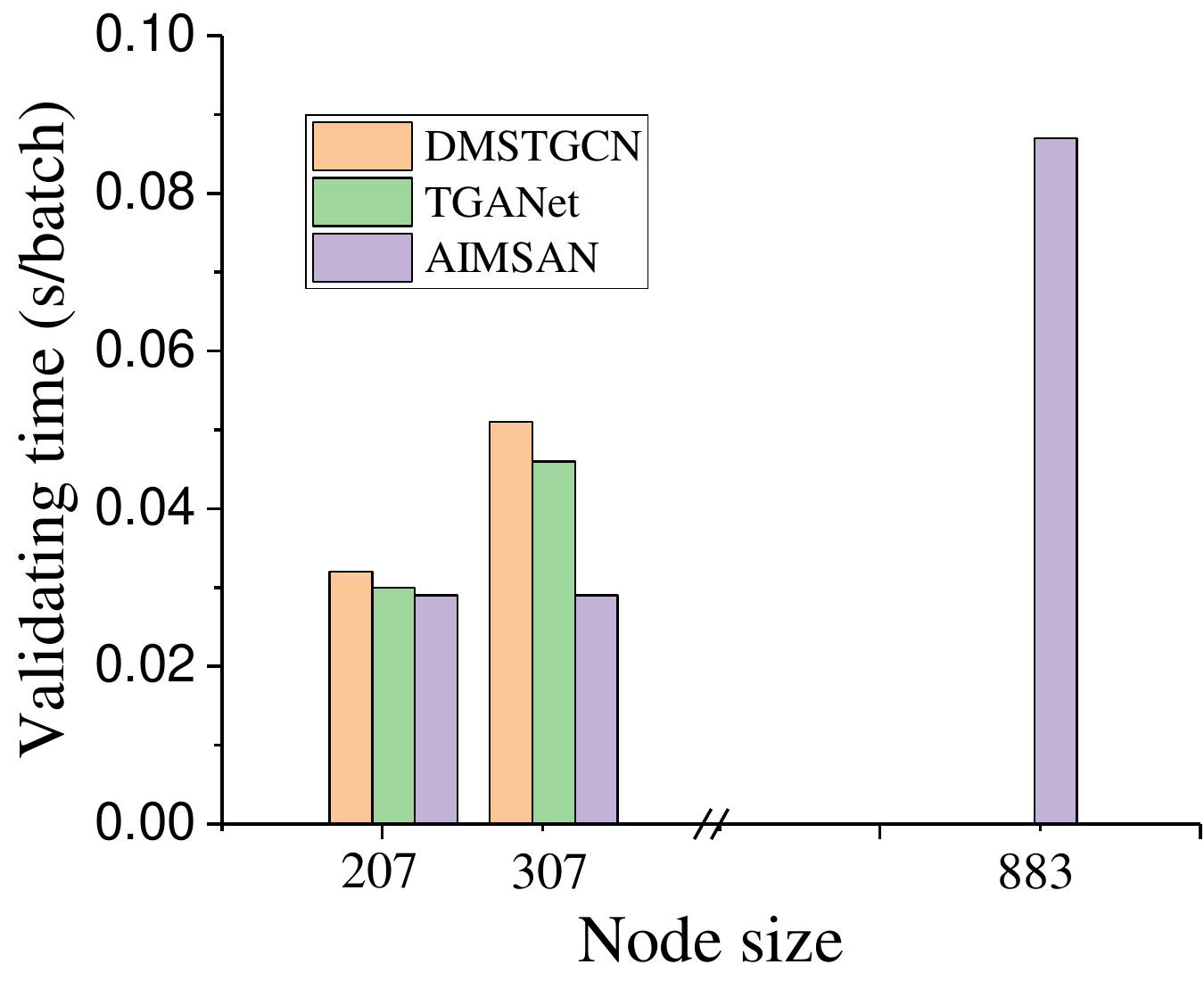}
			\label{fig:valitime}
		}
		\caption{The maximum GPU memory usage, training time and validating time 
		of three methods}
		\label{fig:meo_train_valid_time}       
	\end{figure*}
\end{small}

ST-MetaNet and AIMSAN are two meta-learning-based methods. The differences 
between them are as follows: 1) AIMSAN considers not only the positional 
information of the traffic graph but also the temporal information and weather 
conditions per sample. Therefore, it can learn more hidden temporal-spatial 
information. 2) Except for historical auxiliary information, AIMSAN combines 
future auxiliary information, like time information (holiday) and weather 
forecast. 3) ST-MetaNet averages the geo-graph attributes in the temporal 
dimension and uses averaged attributes in the following meta-knowledge learners, 
which may lose fine-grained information. AIMSAN uses sample-specific auxiliary 
information and tailors the auxiliary information by layer. In other words, the 
higher network layer focuses on more recent auxiliary information. The results in 
Table \ref{tab:comp} show that AIMSAN outperforms ST-MetaNet in most cases, 
especially when the location information is insufficient (in the PEMS04 case). 4) 
ST-MetaNet stacks multiple GRU and GAN layers, and uses a meta learner in each 
sublayer, which results in the out-of-memory error in PEMS07 experiments.

Table \ref{tab:overall_performance} shows the overall performance of AIMSAN, 
DMSTGCN, ST-MetaNet and TGANet  on three datasets for 12-step prediction. 
In the case of single traffic data (METR-LA), AIMSAN achieves an average 
performance improvement of 5.12\% over DMSTGCN. When DMSTGCN has enough traffic 
data (PEMS04) to train its dual-branch structure, the performance of AIMSAN is 
still similar to that of DMSTGCN, with a maximum of 2.63\% performance loss. 
Besides, AIMSAN achieves an average performance improvement of 3.29\% and 0.71\% 
over ST-MetaNet and TGANet, respectively.

\subsubsection{Memory and runtime overhead}
As listed in Table \ref{tab:dataset}, three datasets have different scales of 
traffic nodes. Therefore, comparing various metrics under these datasets can 
demonstrate the scalability of algorithms toward different scales of traffic 
prediction tasks. Thus, we compare the three best baselines (DMSTGCN, ST-MetaNet 
and TGANet) with AIMSAN. For fairness, we set the batch size to 64, as in DMSTGCN 
\cite{DMSTGCN}, to get the memory and runtime costs.   
For better visualization, we draw the memory usage and runtime under different 
node scales in Fig. \ref{fig:meo_train_valid_time}, where the memory usage and 
runtime of AIMSAN grows much more slowly than its three peers. Hence, it can be 
applied to larger traffic prediction tasks.

\subsection{Parameter sensitivity and Ablation study}
In this subsection, we conduct the parameter sensitivity and ablation study on 
METR-LA.

\subsubsection{Different multi-head sizes}
The multi-head size of an attention module is an essential hyper-parameter in the 
proposed method. Therefore, we discuss the impact of head size on AIMSAN. The 
quantity of computation of multi-head attention increases with the head size, 
ranging from 1 to 8. Therefore, this work does not consider larger head sizes 
(\begin{small}$h> 8$\end{small}). From Fig. \ref{fig:multi-head}, when head size 
\begin{small}$h=3$\end{small}, it has the best performance.

\begin{small}
	\begin{figure}[htbp]
		\centering
		\includegraphics[width=0.6\linewidth]{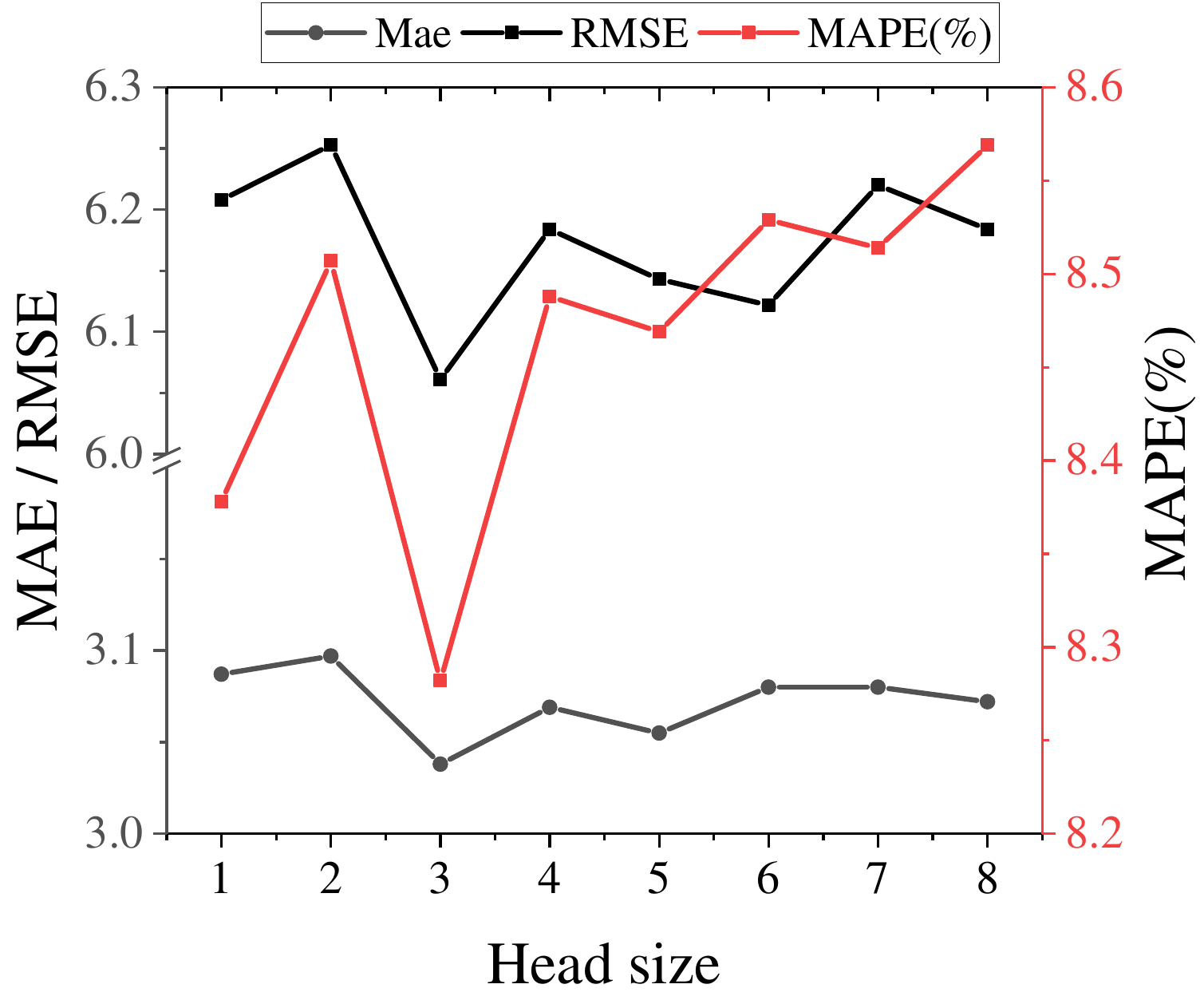} 
		\caption{Performance of AIMSAN under different multi-head sizes.
		}
		\label{fig:multi-head} 
	\end{figure}
\end{small}

\subsubsection{Ablation study of AIMSAN}

\begin{small}
	\begin{figure}[h]
		\centering
		\includegraphics[width=1\linewidth]{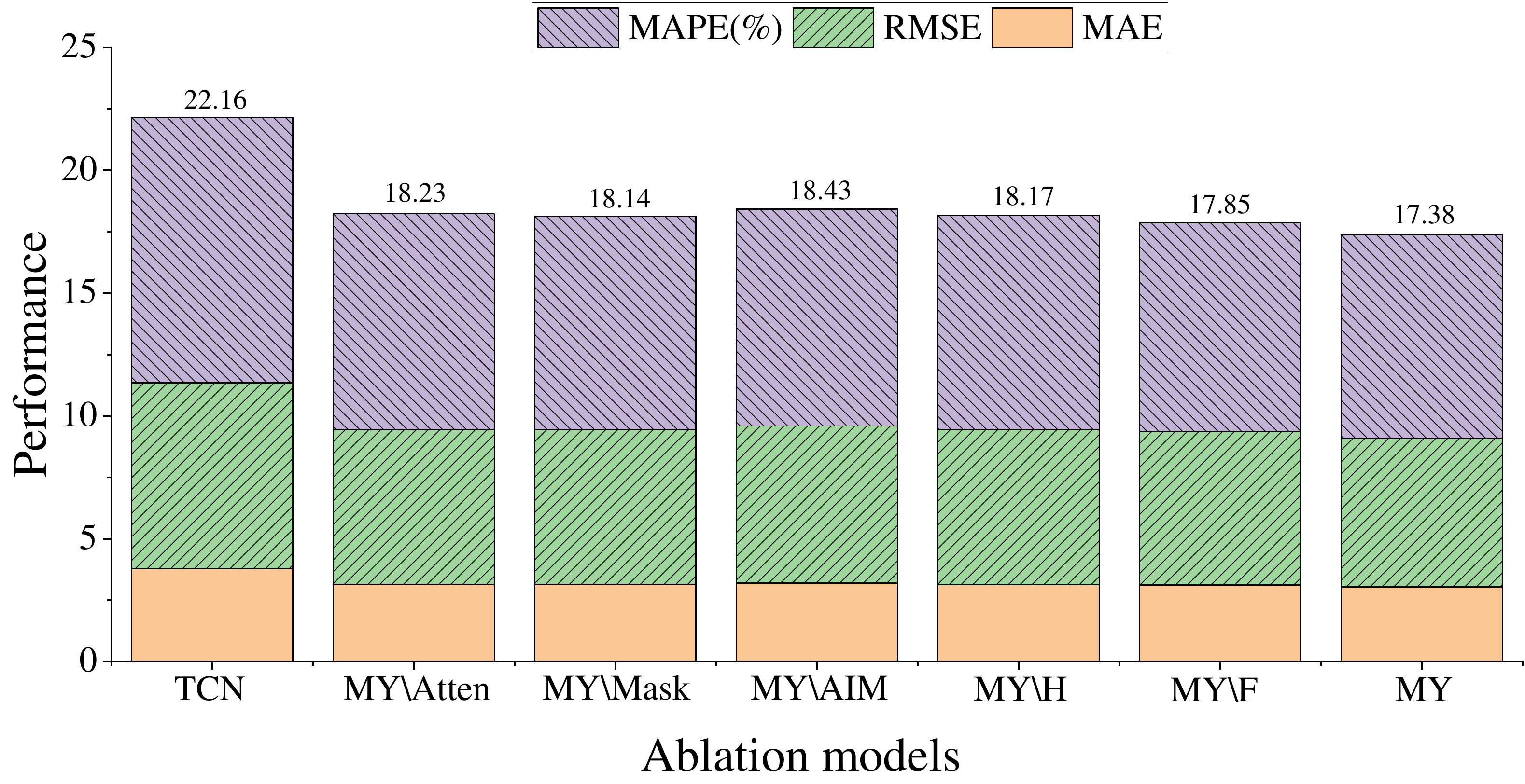} 
		\caption{Ablation study of AIMSAN.
		}
		\label{fig:ablationofmodel} 
	\end{figure}
\end{small}

In this subsection, we study the effectiveness of each module in AIMSAN, and the 
results are illustrated in Fig. \ref{fig:ablationofmodel}. For brevity, we use MY 
to represent the proposed AIMSAN. TCN denotes the simple temporal convolutional 
network in AIMSAN. MY\textbackslash Atten abandons the SAN module in AIMSAN. 
MY\textbackslash Mask cancels the mask operation in the SAN module. 
MY\textbackslash AIM prunes all AIMs in AIMSAN. MY\textbackslash H and 
MY\textbackslash F cancel AIMs for historical and future auxiliary data, 
respectively. From the results, we can draw the following conclusions. 1) TCN can 
learn the temporal dependency of traffic data. However, compared with GCN-based 
methods, pure TCN fails to model complex spatial correlation of traffic data, 
which results in its worse performance. 2) The comparison between 
MY\textbackslash Atten and MYshows the efficiency of the proposed SAN module, 
which can learn the dynamic spatial-temporal correlation of traffic data. 
3) The results of MY\textbackslash Mask and MYverify that masking operation can 
reduce the computational complexity, avoid interference from irrelevant nodes and 
further reduce the prediction error. 4) MY\textbackslash AIM, MY\textbackslash H, 
and MY\textbackslash F show the positive effects of AIM.

\subsubsection{Ablation study on AIM}

\begin{small}
	\begin{figure}[t]
		\centering
		\includegraphics[width=0.8\linewidth]{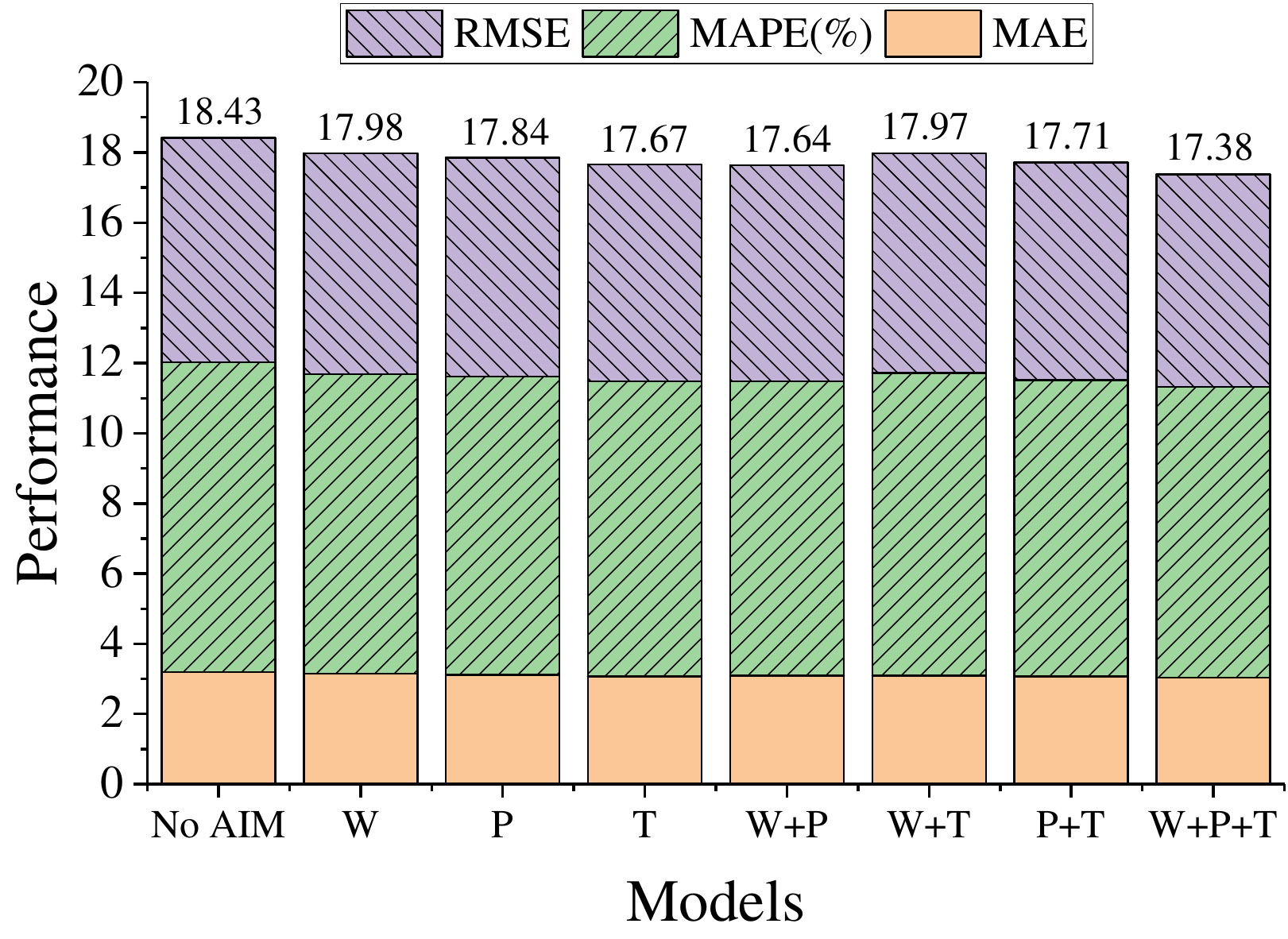} 
		\caption{Ablation study for AIM.
		}
		\label{fig:ablationMK} 
	\end{figure}
\end{small}

AIM  plays an essential role in AIMSAN, which contains weather (W), time (T), and 
position (P) information as shown in Fig. \ref{fig:aip}. Therefore, we further 
make an ablation study on attributes used in AIM. For brevity, we label different 
types of AIMSAN according to the attributes used. For example, W+P+T denotes that 
AIMSAN uses weather, position and time information in the AIM, and W+P denotes 
that AIMSAN only uses weather and position information. From the results, we can 
draw the following conclusion. 1) Weather, position, and temporal information can 
be used to mine hidden correlations within traffic data. 2) Within multiple 
single-attribute cases, the time attribute is superior to the other two. 3) 
Considering multi-attribute auxiliary factors, we obtain the best result using 
all three attributes.

\section{Conclusion}
This paper presents a deep encoder-decoder method (AIMSAN) for traffic data 
prediction, mainly integrating AIM and SAN modules. The former can learn the 
hidden feature of auxiliary information for traffic data, like weather, time and 
position information. In practice, we enrich the weather information of METR-LA 
according to the latitude and longitude of the traffic nodes. Besides, we add 
temporal information like holiday information to three datasets. The latter is a 
sparse cross-attention-based graph convolution module. AIMSAN sets the weights of 
a dynamic adjacent matrix according to the cross-attention values obtained by 
using traffic data and historical auxiliary information. Besides, the diffusion 
graph convolution is used in each SAN module to obtain more spatial information, 
whose diffusion step is 2. Extensive experiments on three public traffic datasets 
(PEMS04, PEMS07 and METR-LA) demonstrate that when considering multiple metrics 
(MAE, MSE, MAPE, memory usage and runtime), the proposed method outperforms its 
peers in most cases. Specifically, AIMSAN has competitive performance with the 
state-of-the-art algorithms but saves 35.74\% of GPU memory usage, 42.25\% of 
training time, and 45.51\% of validation time on average.

\ifCLASSOPTIONcompsoc
\section*{Acknowledgments}
\else
\section*{Acknowledgment}
\fi

This work is funded in part by the National Natural Science Foundation of China 
(File no. 62072216) and the Science and Technology Development Fund, Macau SAR 
(File no. 0076/2022/A2 and 0008/2022/AGJ).

\ifCLASSOPTIONcaptionsoff
  \newpage
\fi

\bibliographystyle{IEEEtran}
\bibliography{AIMSAN.bib}

\begin{IEEEbiography}[{\includegraphics[width=1in,height=1.25in,clip,keepaspectratio]{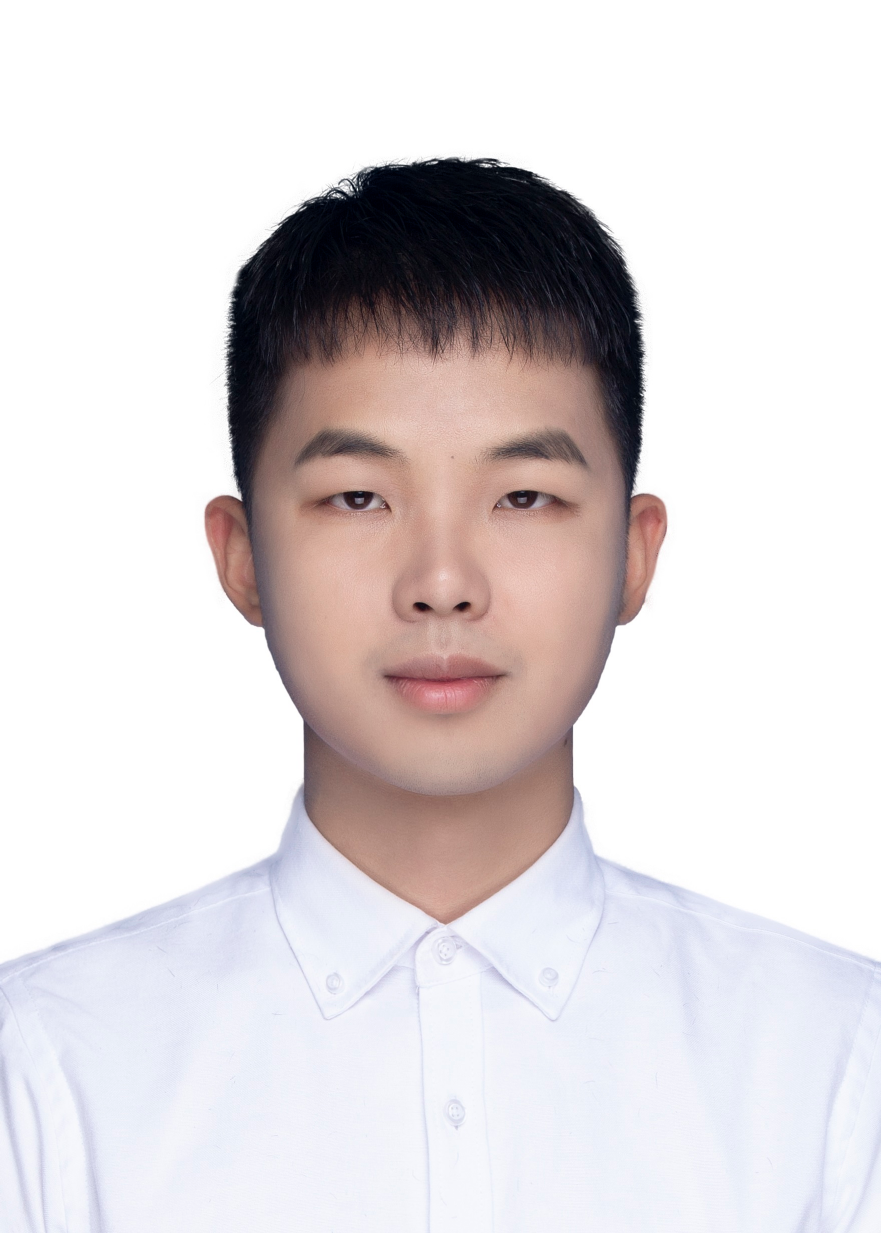}}]{Lingqiang
 Chen}
	received his Ph.D. degree in control science and engineering from Jiangnan 
	University, Wuxi, China, in 2023. He is currently a Lecturer with the School 
	of Information \& Electrical Engineering, Hebei University of Engineering, 
	Handan, China. His research interests are anomaly detection and streaming 
	data prediction on the Internet of Things. He has published several related 
	papers on Inf. Sci., IEEE Trans Instrum Meas, IEEE IoTJ, Appl Soft Comput, 
	ComCom, and Neural Comput Appl.
\end{IEEEbiography}

\begin{IEEEbiography}[{\includegraphics[width=1in,height=1.25in,clip,keepaspectratio]{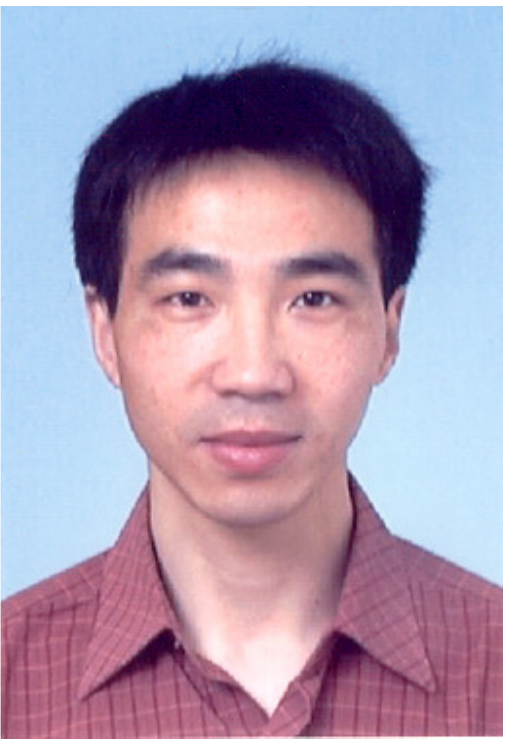}}]{Qinglin
 Zhao}
	received his Ph.D. degree from the Institute of Computing Technology, the 
	Chinese Academy of Sciences, Beijing, China, in 2005. From May 2005 to August 
	2009, he worked as a postdoctoral researcher at the Chinese University of 
	Hong Kong and the Hong Kong University of Science and Technology. Since 
	September 2009, he has  been with the School of Computer Science and 
	Engineering at Macau University of Science and Technology and now he is a 
	professor. He serves as an associate editor of IEEE Transactions on Mobile 
	Computing and IET Communications. His research interests include blockchain 
	and decentralization computing, machine learning and its applications, 
	Internet of Things, wireless communications and networking, cloud/fog 
	computing, software-defined wireless networking. 
\end{IEEEbiography}

\begin{IEEEbiography}[{\includegraphics[width=1in,height=1.25in,clip,keepaspectratio]{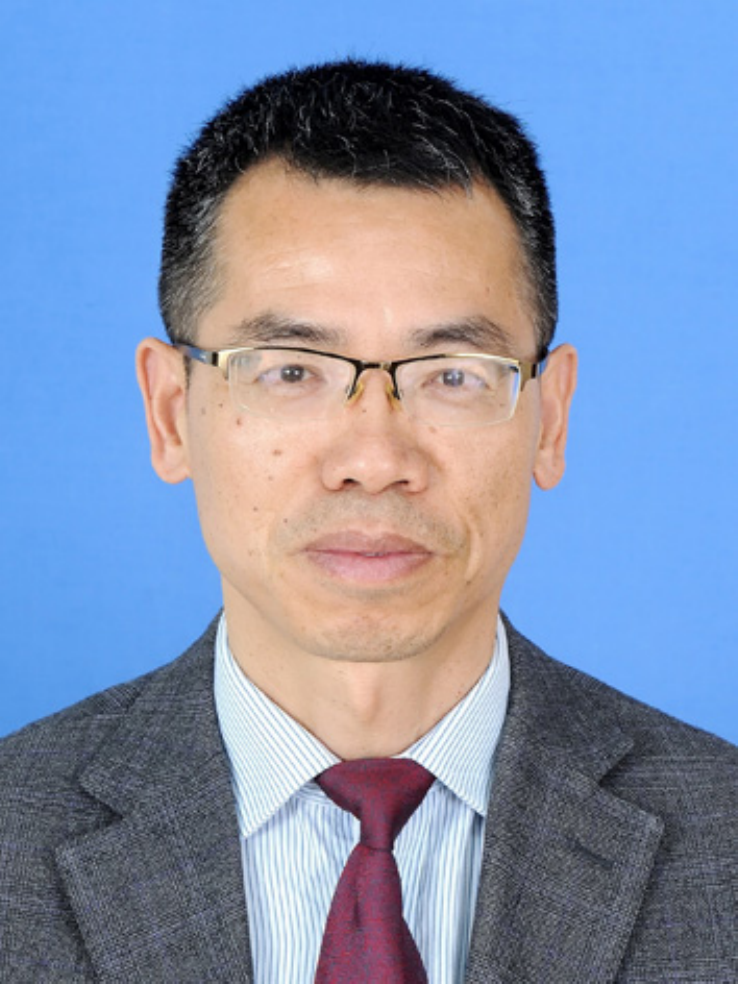}}]{Guanghui
 Li}
	received his PhD degree in Computer Science from the Institute of Computing 
	Technology, Chinese Academy of Sciences, Beijing, China, in 2005. He is 
	currently 
	a professor in the School of Artificial Intelligence and Computer Science, 
	Jiangnan University, Wuxi, China. He has published over 90 papers in journals 
	or conferences. His research interests include Internet of Things, edge 
	computing, fault tolerant computing, and nondestructive testing and 
	evaluation.
\end{IEEEbiography}

\begin{IEEEbiography}[{\includegraphics[width=1in,height=1.25in,clip,keepaspectratio]{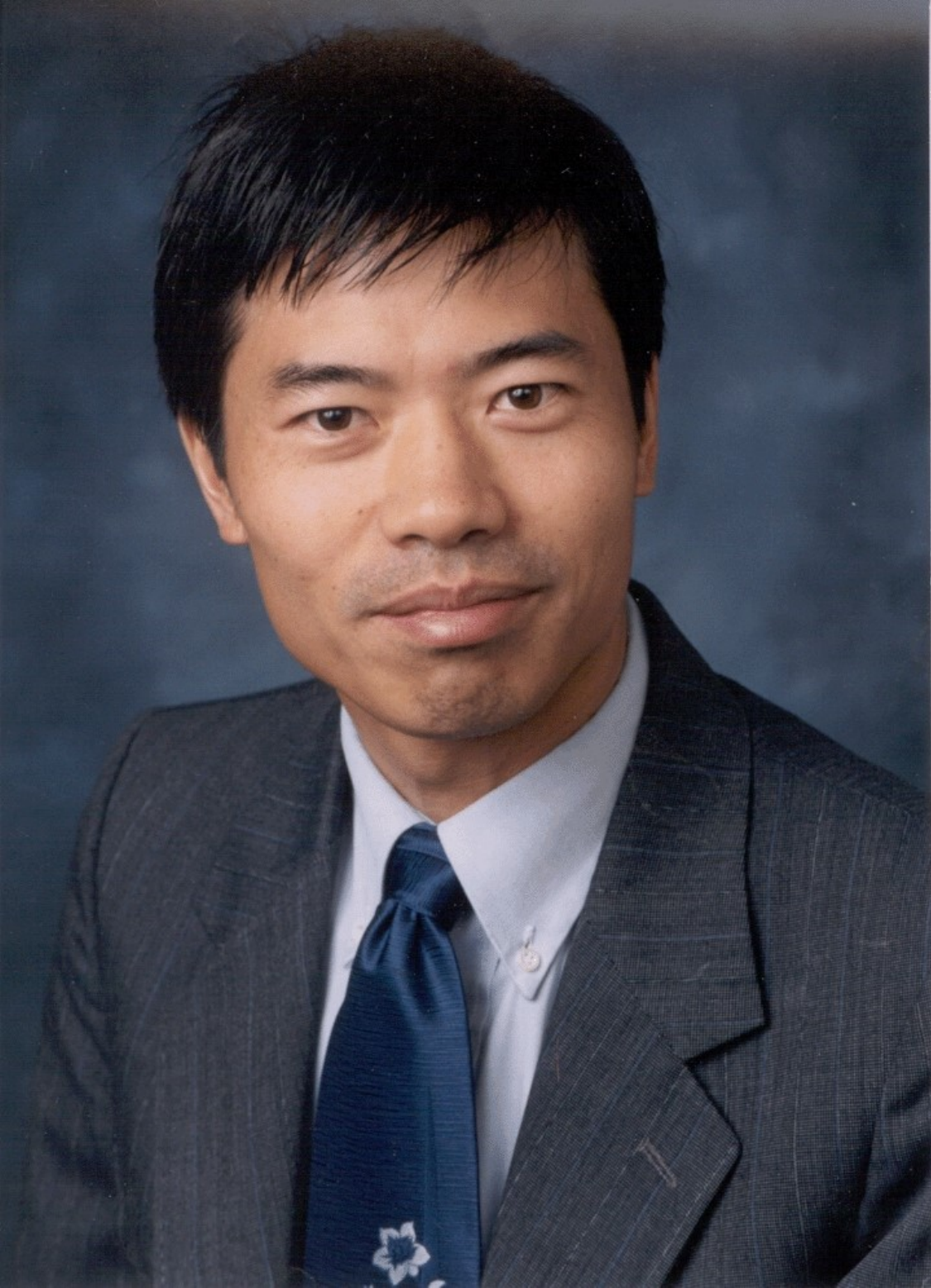}}]
	{Mengchu Zhou}
	(Fellow, IEEE) received his Ph. D. degree from Rensselaer Polytechnic 
	Institute, Troy, NY in 1990 and then joined New Jersey Institute of 
	Technology where he is now a Distinguished Professor. His interests are in 
	Petri nets, automation, robotics, big data, Internet of Things, cloud/edge 
	computing, and AI.  He has 1100+ publications including 14 books, 750+ 
	journal papers (600+ in IEEE transactions), 31 patents and 32 book-chapters. 
	He is Fellow of IFAC, AAAS, CAA and NAI.
\end{IEEEbiography}

\begin{IEEEbiography}[{\includegraphics[width=1in,height=1.28in,clip,keepaspectratio]{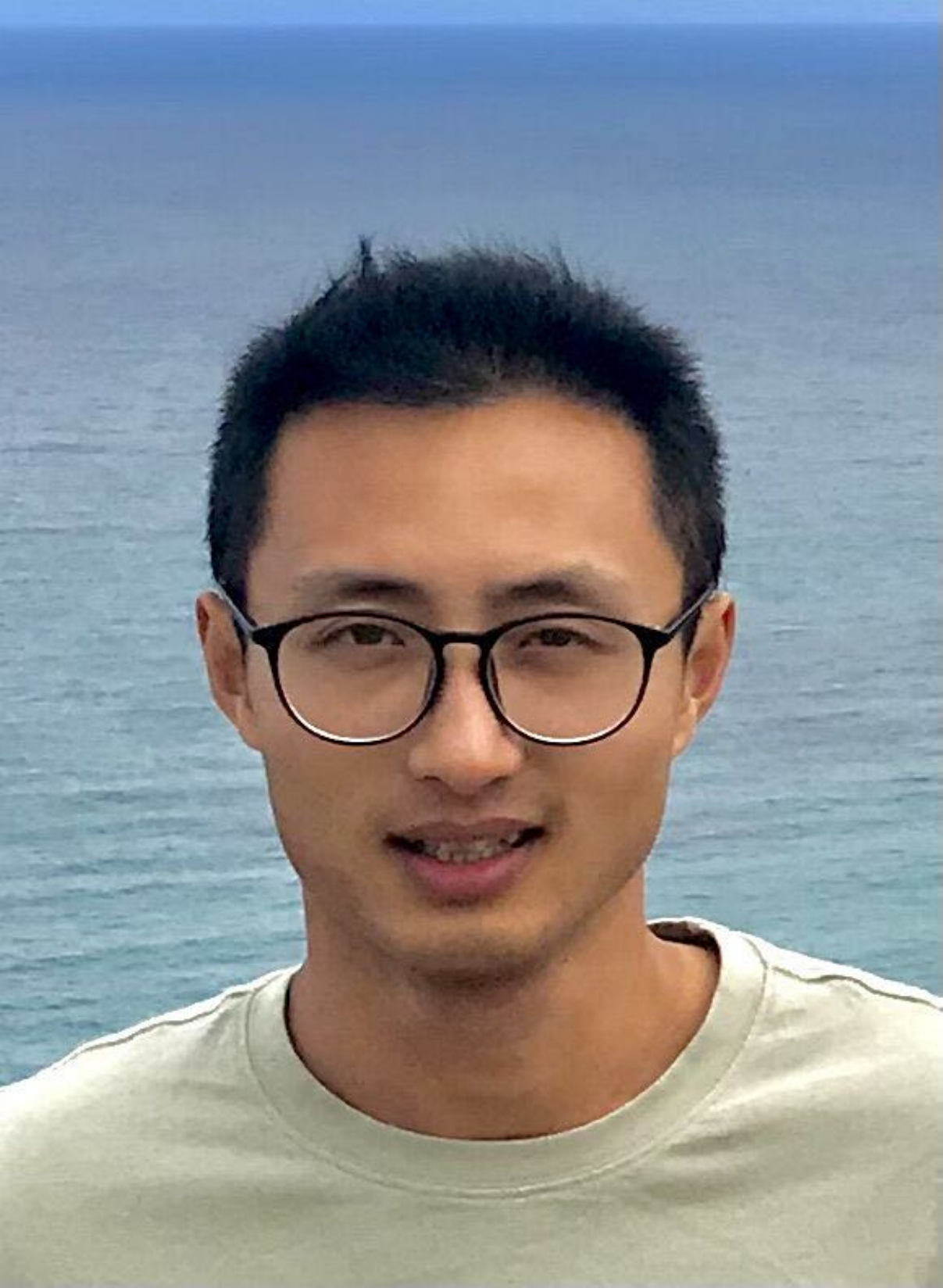}}]{Chenglong
 Dai}
	received the MS, PhD degree in computer science and technology from Nanjing 
	University of Aeronautics and Astronautics, Nanjing, China, in 2014, 2020, 
	respectively. He is currently a Lecturer at the School of Artificial 
	Intelligence and Computer Science, Jiangnan University, Wuxi, China. His 
	research interests include Brain-Computer Interfaces (BCIs), EEG data mining, 
	and machine learning. He has published several related papers in prestigious 
	journals and top conferences, including IEEE TKDE, IEEE TCYB, ACM TKDD, ACM 
	TIST, and SIAM SDM (awarded the Best Paper Award in Data Science Track). He 
	also has served as a Recognition Reviewer for Knowledge-Based Systems, and 
	for IJCNN 18-22.
\end{IEEEbiography}

\begin{IEEEbiography}[{\includegraphics[width=1in,height=1.28in,clip,keepaspectratio]{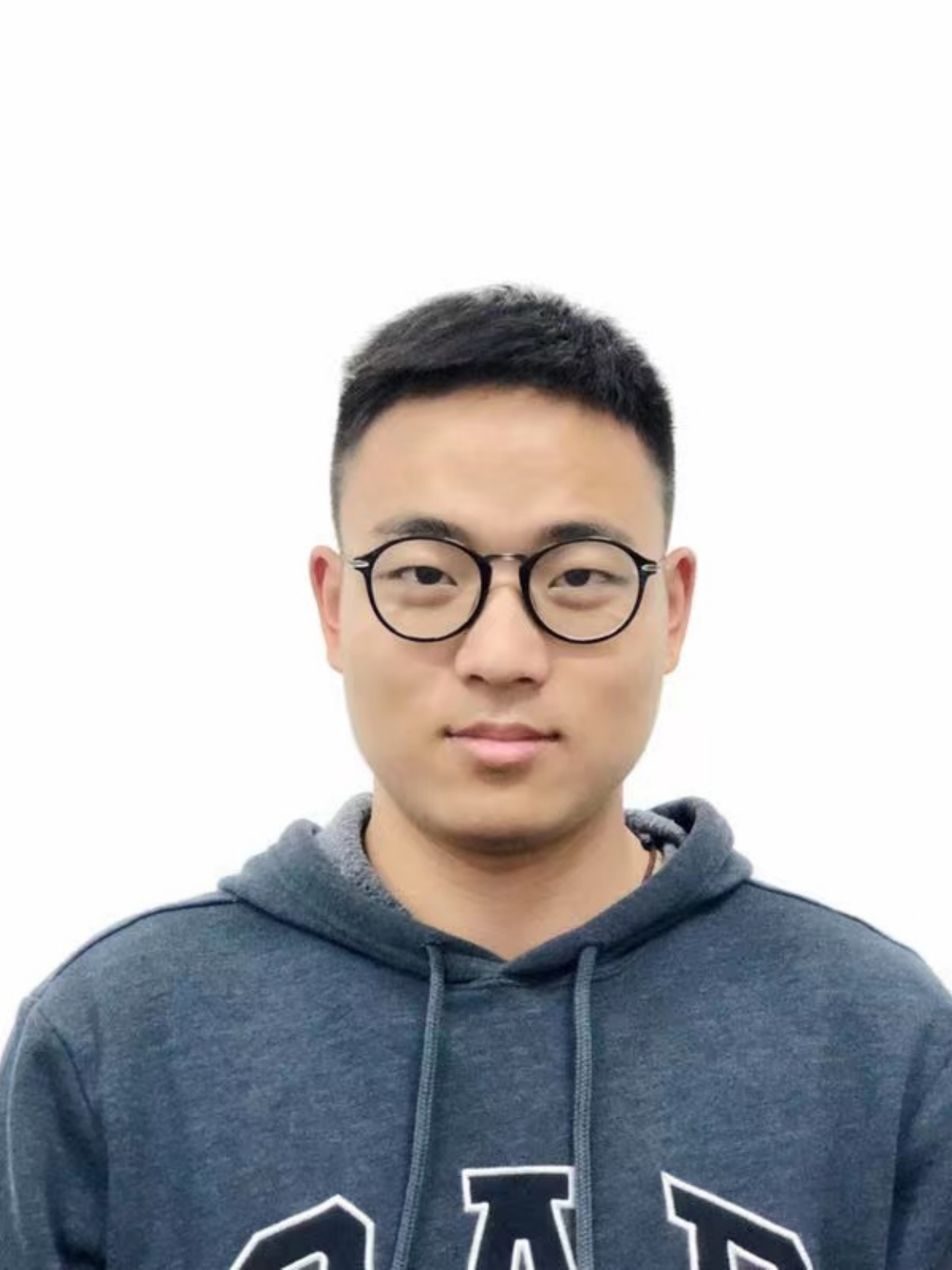}}]{Yiming
 Feng}
	received the B.S. degree in internet of things engineering from Chongqiong 
	University of Technology, Chongqiong, China, in 2018. He is currently a PhD 
	candidate in the School of Artificial Intelligence and Computer Science, 
	Jiangnan University. His current research interests include in edge computing 
	and neural network acceleration.
\end{IEEEbiography}

\end{document}